\documentclass[preprint]{elsarticle}
\pdfoutput = 1
\usepackage[utf8]{inputenc}
\usepackage[a4paper]{geometry}
\usepackage[english]{babel}
\usepackage{hyperref}
\hypersetup{colorlinks=true}
\usepackage{booktabs}

\usepackage{verbatim}
\usepackage{lmodern} 
\usepackage[T1]{fontenc}
\usepackage{microtype}

\journal{Neurocomputing}

\begin{document}

\begin{frontmatter}
\title{Meta-survey on outlier and anomaly detection}
\author[1]{Madalina Olteanu}
\ead{Madalina.Olteanu@dauphine.psl.eu}
\author[1]{Fabrice Rossi\corref{cor1}}
\ead{Fabrice.Rossi@dauphine.psl.eu}
\author[2]{Florian Yger}
\ead{Florian.Yger@dauphine.psl.eu}

\cortext[cor1]{Corresponding author}

\affiliation[1]{organization={CEREMADE, CNRS, UMR 7534, Université Paris-Dauphine,
PSL University},
addressline={Place du Maréchal de Lattre de Tassigny},
postcode={75016},
city={Paris}, 
country={France}}
\affiliation[2]{organization={LAMSADE, CNRS, UMR 7243, Université Paris-Dauphine,
PSL University},
addressline={Place du Maréchal de Lattre de Tassigny},
postcode={75016},
city={Paris}, 
country={France}}

\begin{abstract}
  The impact of outliers and anomalies on model estimation and data
  processing is of paramount importance, as evidenced by the extensive
  body of research spanning various fields over several decades:
  thousands of research papers have been published on the subject.  As
  a consequence, numerous reviews, surveys, and textbooks have sought
  to summarize the existing literature, encompassing a wide range of
  methods from both the statistical and data mining
  communities. While these endeavors to organize and summarize the
  research are invaluable, they face inherent challenges due to the
  pervasive nature of outliers and anomalies in all data-intensive
  applications, irrespective of the specific application field or
  scientific discipline. As a result, the resulting collection of
  papers remains voluminous and somewhat heterogeneous.

  To address the need for knowledge organization in this domain, this
  paper implements the first systematic meta-survey of general surveys
  and reviews on outlier and anomaly detection. Employing a classical
  systematic survey approach, the study collects nearly 500 papers
  using two specialized scientific search engines. From this
  comprehensive collection, a subset of 56 papers that claim to be
  general surveys on outlier detection is selected using a snowball
  search technique to enhance field coverage. A meticulous quality
  assessment phase further refines the selection to a subset of 25
  high-quality general surveys. 

  Using this curated collection, the paper investigates the evolution
  of the outlier detection field over a 20-year period, revealing
  emerging themes and methods. Furthermore, an analysis of the surveys
  sheds light on the survey writing practices adopted by scholars from
  different communities who have contributed to this field.
  
  Finally, the paper delves into several topics where consensus has
  emerged from the literature. These include taxonomies of outlier
  types, challenges posed by high-dimensional data, the importance of
  anomaly scores, the impact of learning conditions, difficulties in
  benchmarking, and the significance of neural
  networks. Non-consensual aspects are also discussed, particularly
  the distinction between local and global outliers and the challenges
  in organizing detection methods into meaningful taxonomies.
\end{abstract}
\end{frontmatter}

\section{Introduction}
Real world data-intensive applications are susceptible to the
detrimental effects of noise, outliers and anomalies, which are
commonly observed in large-scale data sets. Anomalies and outliers are
generally defined as observations that deviate in an important way
from other observations, a rather vague and flexible
definition. Despite being the subject of study since the early stages
of modern statistics (see e.g. \cite{Edgeworth1887Discordant}),
effectively dealing with them remains a persistent challenge.

Over the course of several decades, thousands of research papers have
been published on those topics, alongside numerous surveys, review
papers and textbooks. Monographs focusing on statistical treatments of
outliers can be traced back at least to Hawkings' seminal book
\cite{Hawkings1980IdentificationOutliers} while survey articles have
spanned almost two decades, with some early works appearing in 2003
\cite{MarkouSingh2003NoveltyDetectionStatistical}. The richness of
this survey literature has two significant implications.

Firstly, it poses difficulties for researchers in selecting relevant
surveys to read and gauging their content. For instance, questions may
arise regarding the continued relevance of a paper published two
decades ago, as well as the expected background knowledge required
when perusing a recent survey. 

Secondly, this vast literature provides
a unique perspective on the historical evolution of the outlier concept
and  the diverse approaches adopted by researchers from various
communities, including database specialists, computer engineers,
statisticians, machine learning experts, and more.

The objective of this paper is to analyze research papers that claim
to be general reviews on anomaly and outlier detection. By doing so,
it aims to address the questions raised by the extensive and diverse
literature in this field, offering recommendations for reading and
providing a historical perspective. To achieve this goal, the paper
undertakes the first systematic meta-survey on anomaly and outlier
detection, specifically focusing on survey papers rather than standard
research papers (hence the \emph{meta} aspect of the survey). To
ensure the avoidance of selection bias, a classical systematic survey
protocol, as outlined in \cite{Kitchenham2004ProceduresPerforming}, is
followed.

Based on pilot study presented at the ESANN conference in 2022
\cite{OlteanuRossiEtAl2022ChallengesAnomaly}, the meta-survey
begins with a paper collection phase. Two specialized search engines
are utilized to identify a substantial collection of nearly 500 papers
related to outlier and anomaly detection. Through a manual analysis, a
subset of 56 papers is carefully selected based on a strict definition
of a "general survey on outlier and anomaly detection." Subsequently,
a quality assessment is conducted, uncovering unexpected instances of
plagiarism and leading to the refinement of the collection to 25
high-quality papers.

As an additional outcome of the paper collection process, a brief
discussion is provided on specialized surveys. This discussion
primarily focuses on the types of methods, application fields, and
other factors that have been deemed sufficiently important or active
to warrant the effort of reviewing corresponding sections of the
literature. 

Armed with this exhaustive collection of surveys, we can address
several research questions.  The first set of inquiries pertains to
the field as a whole and encompasses historical aspects,
methodological considerations, and paper structure. Specifically, we
aim to assess the extent to which surveys are conducted in a
systematic manner and how their findings are organized. We also delve
into the interplay between surveys, examining aspects such as
integration (e.g., citation of previous surveys) and the emergence of
communities, both in terms of vocabulary and citations. Of particular
interest is the temporal component of this integration, shedding light
on how the field has evolved over time.

A second series of questions can be elucidated by examining in more
details the content of the surveys themselves. We begin by discussing
a selection of consensual topics, emphasizing the process of
consolidating knowledge within these areas. For example, we explore
how the definition and categorization of outliers have undergone
increasing refinement over time. Additionally, we investigate the
enduring presence of artificial neural networks in the field,
particularly the transition from "shallow models" to the advent of
"deep learning." The final research focus centers on non-consensual
aspects, exploring topics where the literature presents divergent
viewpoints on common issues.

By addressing these research questions, we aim to provide insights
into the field of anomaly and outlier detection, shedding light on its
historical development, knowledge consolidation, and areas of
disagreement.

The paper is organised as follows. Section
\ref{sec:outliers-anomalies} recalls some important concepts about
outliers and anomalies, in a historical perspective. It provides a
background for the systematic meta-survey. Section
\ref{sec:methodology} describes in details the meta-survey methodology
and its implementation. Section \ref{sec:glob-analys-select} provides
a global high level analysis of the selected surveys. Section
\ref{sec:consensus} discusses the main consensual findings that can be
gathered in from the surveys. Finally Section \ref{sec:debates} is
dedicated to the debated aspects for which different conflicting
visions can be identified in the literature.

\section{Outliers and anomalies}\label{sec:outliers-anomalies}
We discuss in this section three important aspects of the literature
on outlier detection, focusing on historical aspects as well as on the
Aggarwal's monograph \cite{Aggarwal2017OutlierAnalysis}. This provides
an important context for sections \ref{sec:consensus} and
\ref{sec:debates}. We discuss first general definitions of outliers
and anomalies (Section \ref{sec:elements-style}). Then we explain the
emphasis of classical statistical analysis on outlier removal and
robust statistics (Section \ref{sec:estim-under-cont}). Finally, we
present the idea that data models are the substance of outlier
detection, even when they are implicit (Section
\ref{sec:the-data-model}). Notice that this Section is partially
informed by the pilot study conducted in
\cite{OlteanuRossiEtAl2022ChallengesAnomaly} and by the publications
collected as part of the meta-survey, see Section
\ref{sec:methodology} for details.

\subsection{Elements of style}\label{sec:elements-style}
Historically, anomalies and outliers have been defined in plain
English using arguably vague sentences such as 
\begin{itemize}
 \item an anomaly ``\emph{appears to deviate markedly from other
        members of the sample in which it occurs}''
      \cite{Grubbs1969ProceduresDetecting}; 
 \item an outlier is ``\emph{an observation which deviates so much
        from the other observations as to arouse suspicions that it
        was generated by a different mechanism}''~\cite{Hawkings1980IdentificationOutliers}; 
 \item an outlier is ``\emph{an observation (or subset of observations)
     which appears to be inconsistent with the remainder of that set
     of data}'' \cite{BarnettLewis1978OutliersStatistical}. 
 \end{itemize}
Those informal definitions were used as guiding principles to build
mathematically sound and operational definitions. Indeed the ground
work on outlier detection was conducted by statisticians and a natural
translation of the above principles involves hypotheses on the
distribution of the data.

For instance Hawkins distinguishes in
\cite{Hawkings1980IdentificationOutliers} two types of outliers: (i)
either coming from the extreme cases of an heavy tailed distribution
or (ii) from a ``contaminant'' distribution in a two
distributions generating process, where one main distribution
generates the ``good'' observations. A similar distinction is proposed
in \cite{BeckmanCook1983Outliers} with different names: the discordant
observation is either a ``natural variation'' (as in (i)) and the
model has weaknesses which corresponds to type (ii). 

While it is quite common in the literature to use interchangeably
anomaly and outlier (and in some papers, novelty), the models
discussed above hint at a possible distinction. In the first case, an
outlier is compatible with the chosen data distribution, albeit very
uncommon. In the second case, outliers obey to a different
distribution than the normal instances. While they still may be
generated by the normal distribution, those outliers are in some sense
``more abnormal'' than in the first case. Thus following,
e.g. Aggarwal in \cite{Aggarwal2017OutlierAnalysis}, we may use the
term outliers for large deviations and anomalies for extreme ones (we
can even distinguishing weak outliers from strong outliers, if we want
to distinguish between rare and very rare cases). 

It should also be noted that the distinction is somewhat
historical. Older text books such as
\cite{Hawkings1980IdentificationOutliers,BarnettLewis1978OutliersStatistical,Huber1981RobustStatistics,RousseeuwLeroy1987RobustRegression}
use almost exclusively the term outlier (for instance
\cite{BarnettLewis1978OutliersStatistical} has
approximately 3300 occurrences of outliers versus 47 uses of
anomaly). \cite{Aggarwal2017OutlierAnalysis} is more balanced with
roughly 4 times more use of outliers than anomalies, but some recent
surveys such as
\cite{SamariyaThakkar2021ComprehensiveSurvey,RuffKauffmannEtAl2021UnifyingReview,Foorthuis2021NatureTypes}
use an inverse balance. 

\subsection{Estimation under contamination}\label{sec:estim-under-cont}
In the pioneering statistical literature on outliers, the focus is on
\emph{outlier treatment} \cite{Miller1993TutorialReview}, with a 
distinction between \emph{deterministic} outliers and \emph{random}
ones. The first case corresponds to errors in a broad sense. In this case,
the goal is to detect the outlier/error to fix it or reject it. This
is classically done with test of hypothesis.

The second case is more general and corresponds to situations where the
offending observations cannot be explained by some mechanism (and
hence be considered as an error). As in the first case, tests of
hypothesis can be used to determine whether an observation is an
outlier. The outcome includes rejecting the observation, modifying the
assumed data distribution model to make the observation more plausible,
or accommodating outliers in the model.

The common point between the two cases is the need for a model of
normal behavior that is used to build the test, among other
things. The core difficulty is that excepted in very specific cases,
the normal behavior is at best partially known. In particular, under a
parametric model assumption, parameters must be estimated (generally
by maximum likelihood). However, as pointed out in
\cite{HadiImon2009DectectionOutliers}, we are facing a
``chicken-and-egg'' problem: if we knew the true model, it would be
easy to identify outliers, while if we knew which observations have
been generated by the true model (and thus are not outliers), it would
be easy to estimate the parameters of the model (or more generally to
assess the quality of our hypotheses on the data generation model).

This explains why outlier detection has been deeply linked to \emph{robust
statistics}. The goal of this field of statistics is to derive
estimation procedures that are robust to arbitrary errors in the data
\cite{HuberRonchetti2009RobustStatistics,Huber1981RobustStatistics}. Examples
include robust estimations of the covariance matrix of multivariate
data \cite{HubertRousseeuwEtAl2005MultivariateOutlier} and robust
regression \cite{RousseeuwLeroy1987RobustRegression}. We refer the
reader to Rousseeuw and Hubert recent survey on outlier treatment with
robust methods \cite{RousseeuwHubert2018AnomalyDetection} for
details. Notice that even very basic statistical estimators can break
when confronted with a small percentage of outliers. This is the case
of mean and of the standard deviation which are used in a routine way
as a preprocessing step to a lot of methods. There is therefore a lot
of value in using robust approaches even when not directly considering
the case of outlier detection and mitigation. 

The emphasis of early approaches to outlier detection on outlier
treatments can be seen as a manifestation of one of the two cultures
discussed by Breiman in \cite{Breiman2001StatisticalModeling}. The
goal is here to fit properly a model to a data set in order to
analyse the model and interpret it. If a linear regression model is
adequate, the values of its coefficients provide insights on the
influence of the explanatory variables on the target variable. In this
context, outliers are a nuisance. We just want to make sure they do
not constitute a proof of inadequacy of the model to the real world
and that they do not break our estimation procedures. In this end, we
just want to get rid of them! However, in the more general data
science context, the way outliers are handled can be quite
different. For instance in intrusion detection
\cite{PatchaPark2007OverviewAnomaly,TsaiHsuElAl2009IntrusionDetection,Yu2012SurveyAnomaly,BuczakGuven2016SurveyData},
the normal behavior model is not interesting \emph{per se} and serves
only as a detector of intrusions. While the problem of robustness is
still present, the detection performances are more important that the
consistency of parameter estimates, for instance. 

\subsection{Models are everywhere}\label{sec:the-data-model}
As pointed out in \cite{Aggarwal2017OutlierAnalysis} one can summarise
the whole field of anomaly detection as follows:
\begin{quote}\emph{
  Virtually all
  outlier detection algorithms create a model of the normal patterns
  in the data, and then compute an outlier score of a given data point
  on the basis of the deviations from these patterns.}
\end{quote}
This is further summarised by the sentence: ``\emph{the data model is
everything}'' \cite{Aggarwal2017OutlierAnalysis}. According to this
view, the discussion above about the ``chicken-and-egg'' problem
\cite{HadiImon2009DectectionOutliers} of anomaly detection applies to
the whole field, even to methods that do not have an obvious
statistical interpretation. In general, the data model will be
adjusted to the observations \emph{blindly}, that is without knowing
in advance whether a given observation is normal or not. Thus model
fitting must be somehow ``robust'' to the presence of outliers (not
necessarily in the robust statistics sense).

In addition, in the mixture/dual distribution point of view, when one
can hypothesize a model both for the normal data and for the outliers,
the problem of assigning observations to one of the two models is
plagued by the imbalanced nature of the data \cite{HeGarcia2009LearningImbalanced,Krawczyk2016LearningImbalanced,MaHe2013ImbalancedLearning}. By essence anomalies are
rare and thus only simple models can be adjusted to them. 

\section{Methodology}\label{sec:methodology}
This section presents the methodology used to select relevant surveys
to include in the present work. This methodology is based on general
principles of systematic reviews, see
e.g. \cite{Kitchenham2004ProceduresPerforming} and on a pilot study
conducted for the ESANN 2023 conference
\cite{OlteanuRossiEtAl2022ChallengesAnomaly}. 

The paper selection has been conducted
using two specialised search engines, Google
Scholar\footnote{\url{https://scholar.google.com}} (GS) and Semantic
Scholar\footnote{\url{https://www.semanticscholar.org/}} (S2). 
Google Scholar does not provide an API and forbids the use of bots via its
robot.txt file. As such, we use manually the site under the private
mode of the Firefox
browser\footnote{\url{https://www.mozilla.org/en-US/firefox/new/}} to 
avoid potential results tailoring. Semantic scholar was used via its dedicated API. 
The default ordering of both search engines was used. 

\subsection{Meta survey scope}
We are interested in \emph{general} surveys about outlier and anomaly
detection. A publication is considered relevant if it fulfills the
following conditions:
\begin{enumerate}
\item it must be written primarily in English (an abstract in another
  language does not prevent the inclusion into the meta survey);
\item it must discuss a significant number of prior works on anomaly
  detection in an organised way;
\item it must be a peer reviewed article published in a journal, in a
  collection book or in conference proceedings. This excludes
  explicitly submitted papers, technical reports and student oriented
  workshops. This also excludes monographs, text books and
  tutorial. They are used as a way to provide context for the
  meta-survey (some technical reports are also considered for this task);
\item it must be general, that is it should not target only a specific
  subset of the scientific literature based on restrictions such as:
  application field (e.g. computer security), classes of methods
  (e.g. deep learning methods), data types (e.g. anomalies in graphs)
  or learning conditions (e.g. streaming data). We made an exception
  by considering that categorical data and high dimensional data are
  so pervasive that surveys considering only this type of data are
  general enough to be included. 
\end{enumerate}

\subsection{Initial queries}
The first paper selection was made using simplistic queries to cover a
large spectrum of papers, favoring recall over precision. We used the
four possible combinations of \emph{outlier} or \emph{anomaly} with
\emph{survey} or \emph{review}. This choice was based on the results
of the pilot study conducted in
\cite{OlteanuRossiEtAl2022ChallengesAnomaly}. During this study, we
observed firstly that historical works as well as textbooks
(e.g. \cite{Aggarwal2017OutlierAnalysis}) systematically use the words
outlier or anomaly, or both, to describe the subject of interest. In
addition we observed that all the general surveys we were aware of
appeared at the top of the results of those queries. Those queries
were therefore considered to be general enough to have a high
recall. This is confirmed by the significant intersection between the
results sets (see below) as well as the limited impact of the snowball
search (see Section \ref{sec:snowball-search}).

On each search engine, we selected the
first 100 papers for each query. On S2, the four queries reported a
total of 374 distinct paper ids (some articles may have multiple ids:
we found 5 duplicated papers). On GS, we obtained 266 distinct paper
``clusters'' (a cluster contains several paper descriptions that refer
to the same paper, but we still identified a duplicated paper). We
identified 141 common papers and a combined collection of 492 unique
papers (over a total of 800 initial results).

We read the title, abstract and in some cases the full paper in order
to classify the 492 candidate documents into the following six
classes:
\begin{enumerate}
\item excluded documents based on ``technical aspects'': non English
  papers (4), non peer-reviewed documents (2), preprints (3), non existent documents
  (2) and a blatant case of plagiarism;
\item excluded documents based on their content: papers whose main
  subject is not outlier detection;
\item excluded documents based on their nature: monographs, text books
  and tutorial about outlier detection;
\item papers about outlier detection but that are not surveys (e.g. description
  of new methods, applications, etc.);
\item survey papers about specific aspects of outlier detection;
\item general survey papers about outlier detection (as claimed by the
  authors of the papers). 
\end{enumerate}
The breakdown of the 492 papers into the six classes is given by Table
\ref{paper:search:results}\footnote{The list of the papers with the chosen
classes are available here
\url{https://github.com/fabrice-rossi/outlier-anomaly-detection}.}. 

\begin{table}[ht]
  \caption{Classification of the 492 papers identified by S2 and GS
    queries into the six classes defined in the main text (those
    figures concern the first phase only and do not include the papers
    obtained by the snowball search).}\label{paper:search:results}
\centering
\begin{tabular}{llr}
  \toprule
 Type & Class & number of papers \\ 
  \midrule
Excluded & C1 &   12 \\ 
Not on outliers & C2 & 184 \\ 
Text book or tutorial & C3 &   10 \\ 
Not a survey & C4 &  69 \\ 
Not a general survey & C5 & 170 \\ 
General survey & C6 &  47 \\ 
   \bottomrule
\end{tabular}
\end{table}

\subsection{Snowball search}\label{sec:snowball-search}
In order to extend the coverage of our selection, we used a classical
snowball search approach: references of 47 papers selected in the
first phase were analysed to find other general survey papers (class
C6). As outlier detection is a long tradition in statistics, some
papers and books were published before 2000. We decided to sort those
publications in a historical group that was used to write Section
\ref{sec:outliers-anomalies}: this gave us some historical background
and a mean to discuss the temporal evolution of the way outliers are considered
(only in a qualitative way as we cannot claim exhaustivity for those
older publications). Finally, we also included text books and monographs
from those references (class C3). After the snowball process, the
class C6 consists in 56 papers: only 9 general survey papers were
found during this part of the search process, which confirms the large
coverage induced by the simplicity of the search queries. 

\subsection{Quality assessment}\label{sec:quality-assessment}
The quality of the papers selected during the queries and the snowball
search is very uneven. There are in particular two major sources of
quality problems. Firstly, plagiarism is non negligible: 13 papers
among the 56 selected ones show different levels of plagiarism as
detailed in Section \ref{sec:plagiarism}. Secondly  some papers are
simply too short and cover a too small selection of papers to bring
new insights on the field, as explained in Section \ref{sec:contribution}. 

\subsubsection{Plagiarism}\label{sec:plagiarism}
To assess plagiarism, we compared figures between papers and used in
addition the SPECTRE embedding
\cite{specter2020cohan} provided by S2. This embedding assigns to each
paper indexed on S2 a vector representation in dimension 768 based on
its title, abstract and references. The embedding of a paper missing
from S2 was computed using the pre-trained model provided by the
authors. We also computed embeddings for all papers based on their
full content rather than only the abstract (leveraging again the
pre-trained model). For each paper, we computed the five closest
earlier papers in class C6 and in a selection of highly cited papers not in
C6, using both embeddings. Then we compared in details each paper with
its possible inspiration sources. 

We found multiple use of figures from previous papers without proper
credits as will as verbatim or almost verbatim use of their
content. Overall, thirteen papers were found to exhibit various forms
of plagiarism.

More precisely, we identified three papers published after 2009 that
include figures ``borrowed'' without credit from the most cited survey
paper to date \cite{ChandolaBanerjeeEtAl2009AnomalyDetection}. In
addition two papers were found to be obvious plagiarisms of
\cite{ChandolaBanerjeeEtAl2009AnomalyDetection} as they copied not
only figures, but also most of the text with minimal editing. We
spotted one instance of self-plagiarism as well as strong resemblance
between two papers of different authors. Many papers include lengthy
``quotes'' from \cite{ChandolaBanerjeeEtAl2009AnomalyDetection} with
no proper attribution. We found also an obvious plagiarism of a lesser
known survey \cite{Xi2008OutlierDetection}, with subsequent articles
borrowing both from \cite{ChandolaBanerjeeEtAl2009AnomalyDetection}
and
\cite{Xi2008OutlierDetection}. \cite{ZhangMeratniaEtAl2007TaxonomyFramework}
is another example of a source of tables and images copied without
proper reference (Notice that
\cite{ZhangMeratniaEtAl2007TaxonomyFramework} is an unpublished
technical report and hence is excluded from C6). Some papers are also
borrowing texts and images from the second most cited survey paper to
date \cite{HodgeAustin2004SurveyOutlier}.

\subsubsection{Contribution}\label{sec:contribution}
Finally, we assessed the contribution of each paper to the
state-of-the-art while taking into account the publication date. We
consider that a general survey paper contributes to the
state-of-the-art (SOTA) if it fulfills at least one of the following
conditions:
\begin{enumerate}
\item it discusses recent papers, published after the previously
  published surveys or missed by them;
\item it addresses an important general problem of outlier detection
  such as anomaly categories or the rising importance of deep
  learning;
\item it provides a new point of view on the literature, e.g. by
  introducing a new taxonomy of methods or by analysing existing
  methods with respect to e.g. their scalability. 
\end{enumerate}
To our surprise, a lot of the papers do not position themselves with
respect to previous surveys, apart from citing some of them. Among 57
papers, only 14 explain explicitly their contribution compared to
previous surveys.

Many papers include also a rather small selection of
papers, most of them being already mentioned in previous
surveys. As a reference, Table \ref{table:earliest} shows some statistics on
the earliest general surveys found in our search (prior 2010). Apart
\cite{Petrovskiy2003OutlierDetection}, all papers discuss more than 60
papers with a peak at 361 references in
\cite{ChandolaBanerjeeEtAl2009AnomalyDetection} (which was published
in 2009). Those numbers set the bar quite high in terms of literature
coverage and positioning for papers published after 2009. 

\begin{table}
\centering
\begin{tabular}{lrrr}
  \toprule
paper & number of references & number of words (text) & number of pages \\ 
  \midrule
\cite{MarkouSingh2003NoveltyDetectionNeural} & 91 & 15,455 & 23 \\ 
  \cite{MarkouSingh2003NoveltyDetectionStatistical} & 64 & 11,129 & 17 \\ 
  \cite{Petrovskiy2003OutlierDetection} & 28 & 7,843 & 10 \\ 
  \cite{HodgeAustin2004SurveyOutlier} & 66 & 14,827 & 45 \\ 
 \cite{ BenGal2005OutlierDetection} & 67 & 5,089 & 16 \\ 
  \cite{AgyemangBarkerEtAl2006ComprehensiveSurvey} & 80 & 8,743 & 18 \\ 
  \cite{ChandolaBanerjeeEtAl2009AnomalyDetection} & 361 & 23,478 & 58 \\ 
  \cite{HadiImon2009DectectionOutliers} & 69 & 8,940 & 14 \\ 
   \bottomrule
\end{tabular}
\caption{Statistics on the earliest surveys discussed in the present
  paper: number of reference, number of words in the text (excluding
  references), total number of pages (including references)}
\label{table:earliest}
\end{table}

\begin{figure}
  \centering
  \includegraphics{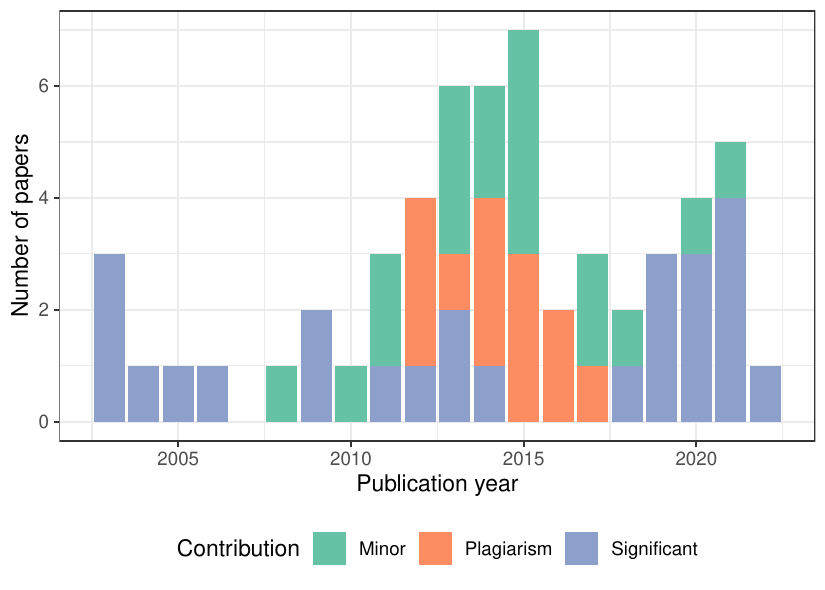}
  \caption{Number of general survey papers published per year sorted
    into papers with plagiarism, papers that have only a minor
    contribution to the SOTA and papers that improve the SOTA.}
  \label{fig:papers_per_year}
\end{figure}

Considering all the criteria described above, we identified only 25
papers\footnote{A summary of our evaluation of the 56 papers in C6 is
  available here \url{https://github.com/fabrice-rossi/outlier-anomaly-detection}.} (among the remaining 44 papers that do not involve plagiarism)
as contributing to the state-of-the-art (see the full list in
\ref{appendix:paper-list}). One paper
\cite{CousineauChartier2010OutliersDetection}
was excluded as it is written for a specific research community
(medical research): it makes underlying assumptions about the data
production process which limits strongly their generality. Its
inclusion in class C6 is even debatable. Figure
\ref{fig:papers_per_year} shows the number of papers published per
year broken down into three categories.

While our sorting process remains expert based and partially
questionable, it aligns relatively nicely with simple numerical
characteristics of the papers. We represented each paper as a low
dimensional vector using the following characteristics:
\begin{itemize}
\item size: total length (total number
  of characters), total text length (total number of characters
  excluding the references), number of pages;
\item literature coverage: number of references and delay in years
  between the publication year of the survey and the publication year
  of the most recent paper cited;
\item citations: the logarithm of the average number of citations per
  year of the paper since its publication according to Google
  Scholar\footnote{The citation numbers were collected in January
    2023};
\item plagiarism: the Euclidean distance between the embedding of the
  survey and the embedding of its closest neighbor in the earlier
  papers, both using the abstract based embedding and using the full
  paper embedding.
\end{itemize}

\begin{figure}
  \centering
  \includegraphics{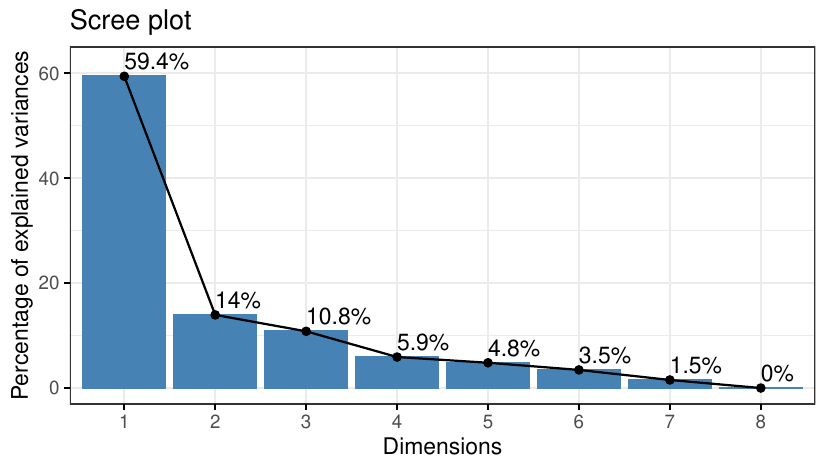}
  \caption{Scree plot of principal component analysis results on the numerical
    characteristics of the survey papers.}
  \label{fig:papers_scree}
\end{figure}

\begin{figure}
  \centering
  \includegraphics{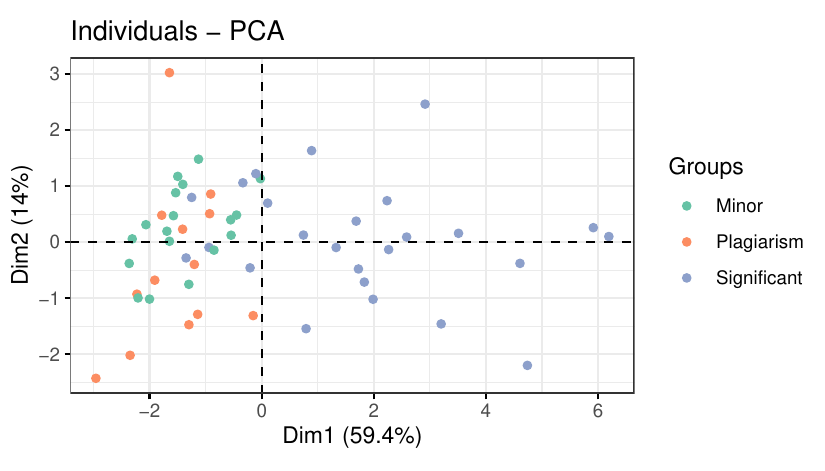}
  \caption{Principal component analysis results on the numerical
    characteristics of the survey papers (first two principal components).}
  \label{fig:papers_pca}
\end{figure}

\begin{figure}
  \centering
  \includegraphics{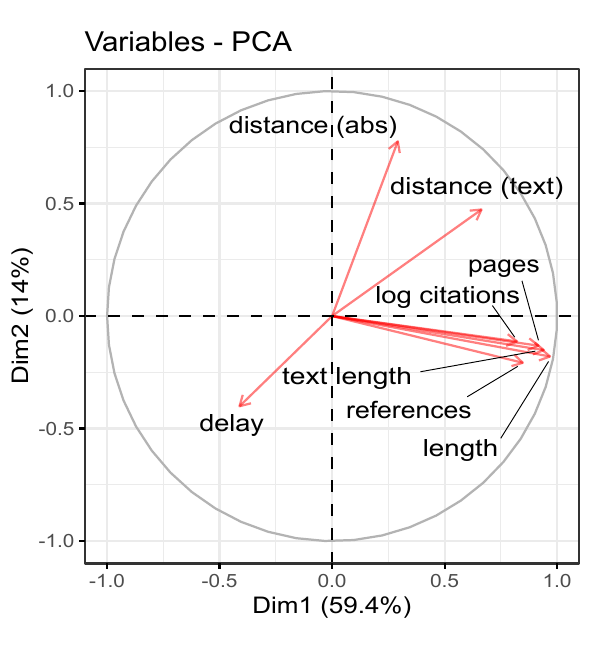}
  \caption{Contributions of the variables to the first two principal
    components of the numerical characteristics of the survey papers.}
  \label{fig:papers_pca_var}
\end{figure}

A simple Principal Component Analysis can be used to capture roughly
73 \% of the variance of the data, as shown in the Scree plot on
Figure \ref{fig:papers_scree}. Figure \ref{fig:papers_pca} shows the
projected papers: most of the papers that contributed to the SOTA have
a high positive value on the first principal axis, while minor papers and
papers with plagiarism have negative values on the same axis. The
second principal axis can be used to separate partial papers with or
with out plagiarism. Figure \ref{fig:papers_pca_var} shows the
contribution of the variables to those axes. The first PC is mostly
explained by size effects, including the number of citations received
by the paper, but also by the freshness of the references. The second
PC is more related to the proximity to other papers. We provide in
\ref{appendix:pca} additional representations of the PCA results in
order to explore the third component (which capture 10.8 \% of the
variance). They confirm the interpretation derived from the first two
components. 

\subsection{Specific surveys}\label{sec:specific-surveys}
To get additional insights on the field, we did some further analysis
of the papers in C5 (surveys on some specific aspects of outlier
detection). Based on the content of the title, the abstract and, if
needed, the full paper, we identified the restriction chosen by
the authors with respect to four axes: application fields, class of
methods, data types and learning conditions. Statistics on the
specificity of those papers are given in Table
\ref{table:restriction}. 

\begin{table}
\centering
\begin{tabular}{lr}
  \toprule
Specificity axis & Number of papers\\
  \midrule
  Application fields &109\\
  Class of methods& 59\\
  Data types&29\\
  Learning conditions& 18\\
   \bottomrule
\end{tabular}
\caption{Number of papers from class C5 that focus on a specific
  aspect of outlier detection, grouped by specificity axe. The total
  is larger than 170 as 45 papers use two axes of specificity.}
\label{table:restriction}
\end{table}

Application fields are quite varied and we identified 39
different expressions used to design them ranging from very broad ones
(such as \emph{computer networks}) to very specific ones (for instance \emph{water
  quality} or \emph{smart grids}), see Table \ref{table:restriction:fields} for the main ones. The field of networked objects
interpreted in a broad sense is largely dominating the application
specific surveys and is itself structured in subfields. Some surveys
cover the general case of networks of computers while others focus on
networks of lower capacity systems such as wireless sensor networks or
internet-of-things (IOT). Some surveys cover more generally intrusion
detection, that is a specific case of anomalies in the context of
networked computers. Finally some surveys cover other specific cases of
security issues such as financial frauds. 

\begin{table}
\centering
\begin{tabular}{lr}
  \toprule
Application field & Number of papers\\
  \midrule
  Computer network (general case) & 22 \\
  Sensor networks & 19 \\
  Intrusion detection & 16\\
  Internet-of-things (IOT) & 9\\
  Financial frauds & 6\\
   \bottomrule
\end{tabular}
\caption{Main application fields of specialized surveys. All other
  fields are specific to three papers at most.}
\label{table:restriction:fields}
\end{table}

Surveys specific to a particular class of methods are essentially
dedicated to deep learning and to a less extent to robust
statistics. We identified indeed 18 surveys focusing on neural networks (among 59 restricted on the
method axis), among which 16 are about deep learning in general, one about Long
Short-Term Memory models (LSTM, \cite{HochreiterSchmidhuber1997LongShort})
and one about Generative Adversarial Network (GAN,
\cite{GoodfellowPougetAbadieEtAl2014GenerativeAdversarial}). Moreover,
among the 16 deep learning surveys, 12 are specific either to an
application field (e.g. IOT) or to a type of data (for instance time
series). We have also identified 7 papers on robust statistics, while
the focus of the remaining 34 papers is more spread from k-means to
clustering.

The data type specialisation axis does not exhibit any dominating
type apart from temporal data which are the main focus of 9 surveys
out of 29. Finally, papers that cover a specific learning context are
mainly dedicated to data streams (14 surveys among 18).

It appears clearly from this analysis that specialized surveys belong
mainly to the field of computer science and computer engineering,
while contributions from statistics seem to be relatively
limited. Another clear outcome is the focus on deep learning which is
sufficient popular to be combined with other restrictions. Those
findings will be confirmed to some extent in the following 
section dedicated to the analysis of the selected papers.

Notice that results of the present section should be considered with
caution and only as building blocks for a systematic survey on the
identified subfields. Indeed because of the search strategy adopted we
may have missed interesting specific surveys on e.g. robust
statistics. Moreover, the analysis has been made on the raw results of
the classification without applying the full methodology developed for
the general surveys (snowball search and quality assessment). Thus the
figures reported here are only indicative. 

\section{Global analysis of the selected survey papers}\label{sec:glob-analys-select}
We discuss in this section general aspects of the selected
papers, in particular their paper collection methodology, or the lack
thereof (Section \ref{subsection:methodology}), their structure (Section
\ref{subsection:structure}), the structure of the field (Section
\ref{sec:structure-field}) and their vocabulary (Section 
\ref{subsection:topics}). 

\subsection{Methodology}\label{subsection:methodology}
It should first be noted that almost none of the surveys include a
proper paper collection methodology. In fact only two of the papers in
class C6 describe the way the papers were collected and selected
\cite{AguinisGottfredsonEtAl2013BestPractice,NassifTalibEtAl2021MachineLearning}.
\cite{CarrenoInzaEtAl2020AnalyzingRare} describes also a systematic
literature collection but in less details and as a way to validate
hypotheses about learning paradigms and their use in research papers.

The absence of a proper paper collection methodology is problematic as
collection bias could be present: for instance the papers discussed in
\cite{SalehiMirzaeiEtAl2022UnifiedSurvey} give the impression that the
contemporary research in outlier detection is almost uniquely
conducted with deep learning approaches, whereas in the slightly older
paper \cite{BoukercheZhengEtAl2020OutlierDetection}, deep learning
papers are a minority.

In addition to those potential biases, the absence of an explicit
paper collection methodology prevents its reproduction and increases
the efforts needed to update the survey in the future. As most surveys
consist in an organised collection of short summaries of selected
papers, their long term value is potentially limited. It seems
therefore important to be able to update them somehow, hence to follow
a proper methodology. 

\subsection{Paper structure}\label{subsection:structure}
Many survey papers are structured in a quite standard way: the authors
identify a collection of interesting papers, arrange them in
categories (potentially structured into a taxonomy) and then provide
for each paper a short summary that contrasts it to other papers in
the same category. In general, high level comparisons between
categories are also provided.

Most of the surveys considered in the present paper do not depart from
this general scheme. In our opinion, this type of structure has more
drawbacks than advantages.

There is of course value in providing a short summary of recent
papers: the number of papers produced each year is enormous and
researchers most focus their attentions to a selection of them. To
illustrate this consider the recent NeurIPS
conferences\footnote{\url{https://neurips.cc/}}. They accepted 2344
papers in 2021 and 2672 papers in 2022. Based on the simple search
tool available on the conference website, we can select papers about
outlier and anomaly detection, 8 in 2021 and 14 in 2022. This would
miss directly related papers that use a slightly different framing,
for instance papers about out-of-distribution detection
\cite{SalehiMirzaeiEtAl2022UnifiedSurvey}. More generally, as the
number of papers published (or simply made available on
arXiv\footnote{\url{https://arxiv.org/}}) keeps increasing,
researchers are likely to miss important papers that are only slightly
departing from their main focus. In this context, summary oriented
surveys can help researchers to remain aware of the progress in the
state-of-the-art on subjects that are closely related to their main
research interests.

However, the value of this type of surveys tend to decrease relatively
quickly. For instance while
\cite{MarkouSingh2003NoveltyDetectionNeural,MarkouSingh2003NoveltyDetectionStatistical}
were very thorough surveys in 2003, their contemporary relevance is mainly
historical and they illustrate by contrast with
e.g. \cite{RuffKauffmannEtAl2021UnifyingReview} the tremendous
evolution of the field in almost 20 years. The recent survey
\cite{SamariyaThakkar2021ComprehensiveSurvey} is currently very useful
as a reference for the algorithmic complexity of a large collection of
methods, but its interest will decrease over the years with the
introduction of new and more efficient methods.

In addition, the organisation of this type of surveys in broad
categories is generally detrimental to the presentation of general
issues (such as the problem induced by high dimensional data, see
Section \ref{sec:high-dimens-issue}). Many of those surveys tend to
get caught into details about the specific algorithms they are
discussing at a given point while missing the point of agreement or
the crucial differences between hypotheses. For instance, as pointed
out in \cite{BoukercheZhengEtAl2020OutlierDetection}, many surveys
distinguish \emph{distance based} and \emph{density based} outlier
detection methods, while they are all essentially based on comparing
distances to nearest neighbors and share therefore a lot of advantages
and limitations.

The need for structure in any paper is fulfilled in this type of work
by relying on categories (and taxonomies) in a way that may seem
somewhat exaggerated. For instance, the idea that there may be a
\emph{statistical} method category is questionable as many
\emph{distance based} approaches can be derived from a Gaussian
assumption or using a form of kernel density estimation. As pointed
out in \cite{ZimekFilzmoser2018ThereBack} most of the \emph{nearest
  neighbors} based approached can be seen as density estimation based
methods, as a consequence as non parametric statistical methods. See
Sections \ref{subsection:outlier:definition} and \ref{sec:taxonomies} for
longer discussions about taxonomies.  

Some surveys considered in the present paper deviate from this generic
structure, partially or completely. In our opinion they are the most
interesting ones from a mid to long term perspective. We will include
their findings in the discussion in Section \ref{sec:consensus}. Their
main contributions in addition to the papers they discussed can be
summarized as follows:
\begin{itemize}
\item they discuss in details the nature of anomalies and outliers
  \cite{AguinisGottfredsonEtAl2013BestPractice,Foorthuis2021NatureTypes}
  by contrasting the different definitions proposed in the literature
  (Section \ref{subsection:outlier:definition}), way beyond the
  traditional separation between point, contextual and collective
  anomalies proposed initially in
  \cite{ChandolaBanerjeeEtAl2009AnomalyDetection};
\item they provide some unifying views on methods that are generally
  discussed independently in summary based reviews
  \cite{Rokhman2019SurveyMixed,RuffKauffmannEtAl2021UnifyingReview,ZimekFilzmoser2018ThereBack};
\item they show links between variants of outlier detection or
  extensions of the concepts, for instance distinguishing rare events
  detection from novelty detection
  \cite{CarrenoInzaEtAl2020AnalyzingRare} or discussing the general
  framework of out-of-distribution detection
  \cite{SalehiMirzaeiEtAl2022UnifiedSurvey};
\item they focus on major general issues such as large scale data
  \cite{ThudumuBranchEtAl2020ComprehensiveSurvey} or high dimensional
  data \cite{ZimekSchubertEtAl2012SurveyUnsupervised} (Section \ref{sec:high-dimens-issue}). 
\end{itemize}

\subsection{Structure of the field}\label{sec:structure-field}
As shown on Figure \ref{fig:papers_per_year}, there is a renewed need
and interest for writing surveys on outlier detection. An initial
series of surveys was published in the early 2000s, followed by a
regular publication of one significant survey every two years. Since
2018, we are witnessing a significant increase in significant survey
publications.

\begin{figure}
  \centering
  \includegraphics{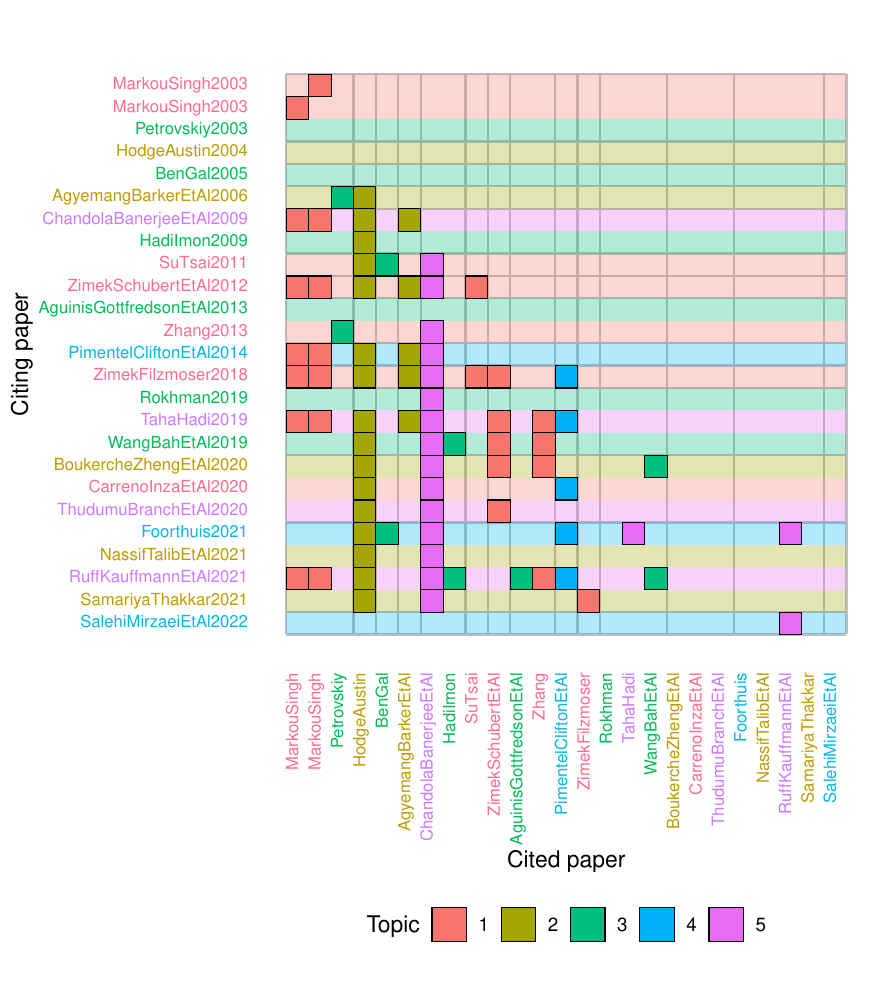}
  \caption{Citations between the survey papers: each row
    (resp. column) represents a paper, ordered by publication year
    (top to bottom, resp. left to right). Thin grey lines separate
    papers by publication years. The color of a paper identifier and
    of the corresponding row corresponds to the topic to which the
    paper is assigned (see Section \ref{subsection:topics}). A colored
    square with a black border on a row shows that the row paper cites
    the column paper.  The filling color is the topic of the cited
    paper. See \ref{appendix:paper-list} for the paper codes.}
  \label{fig:papers_citations_mat}
\end{figure}

The field is relatively integrated in the sense that papers tend to
cite previous surveys, as shown on Figure
\ref{fig:papers_citations_mat}. Highly cited papers, in particular
\cite{HodgeAustin2004SurveyOutlier,ChandolaBanerjeeEtAl2009AnomalyDetection}
are not only cited by general papers but also by almost all the survey
papers collected here.

Nevertheless, we can see that some older papers such as
\cite{Petrovskiy2003OutlierDetection} and
\cite{BenGal2005OutlierDetection} tend to be forgotten. Several recent
papers, such as
\cite{Rokhman2019SurveyMixed,SalehiMirzaeiEtAl2022UnifiedSurvey} tend
to cite only a small selection of recent surveys. This indicates a
potential shift from the first phase of papers published before 2015
to the current phase which started around 2018. Older papers are
replaced by more recent ones mainly as the former lose their summary
oriented value, as discussed in the previous section. See Section
\ref{subsection:topics} for additional remarks based on topic models. 

Another potential source of the drop in important survey publication
around 2015 is the disruption in the machine learning field induce by
the re-emergence of neural networks with the explosion of deep
learning (see Section \ref{sec:neur-netw-models}). With such a
disruption, it takes several years to propose novel approaches in this
new paradigm and then a lag ensues regarding the publication of
surveys. This phenomenon is also potential coupled with high progress
rate of deep learning methods which may explain the large number of
surveys that are both specific to deep learning and to something else
(such as images or IOT) as discussed in Section
\ref{sec:specific-surveys}. 

\subsection{Topic modeling}\label{subsection:topics}

\begin{figure}
  \centering
   \includegraphics{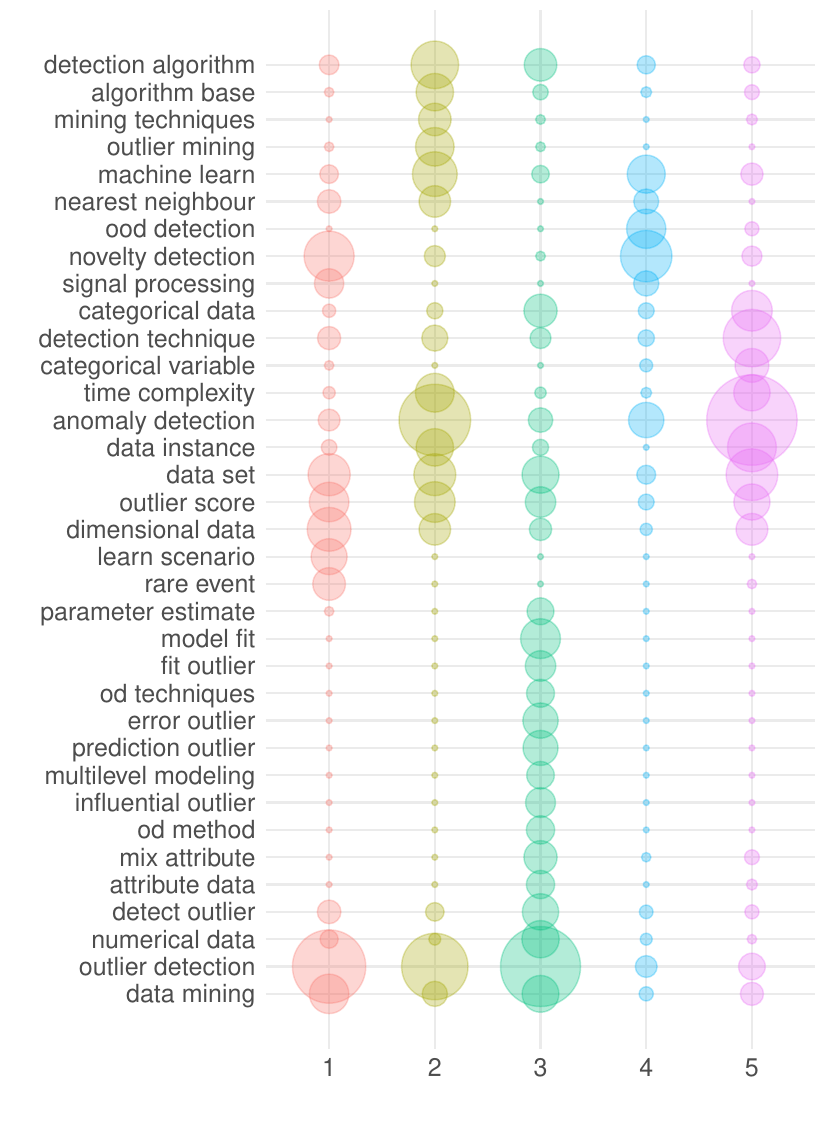}
  \caption{Top 0.1\% of the most salient bigrams in the corpus, for the LDA model. The surface of each disk is proportional to the frequency of the associated bigram within the topic.}
  \label{fig:lda_saliency}
\end{figure}

\begin{figure}
  \centering
\mbox{}\hfill\includegraphics{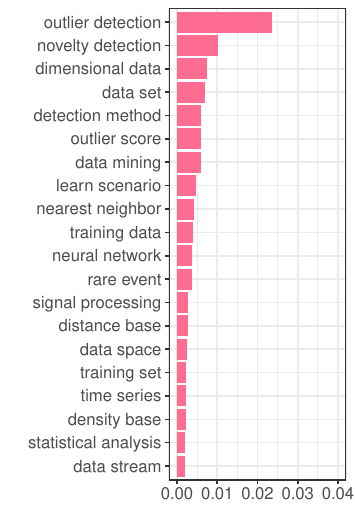}\hfill
    \includegraphics{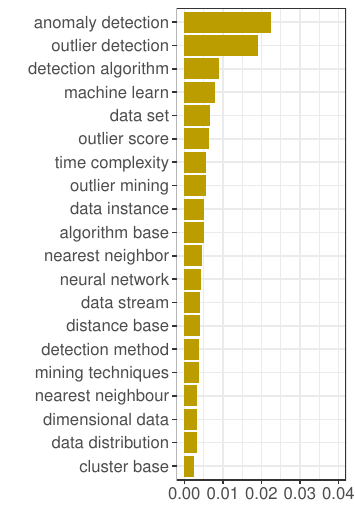}\hfill\mbox{}
  \caption{Most frequent 20 bigrams for topics 1 and 2.}
  \label{fig:lda_freq_1_2}
\end{figure}

\begin{figure}
  \centering
     \mbox{}\hfill\includegraphics{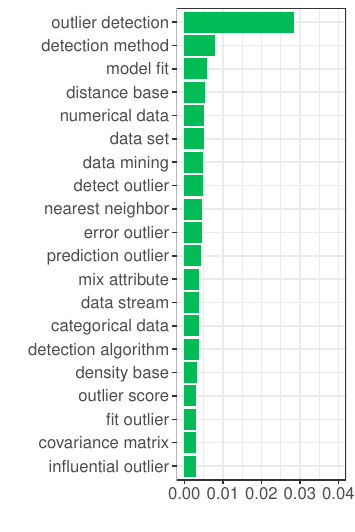}\hfill
     \includegraphics{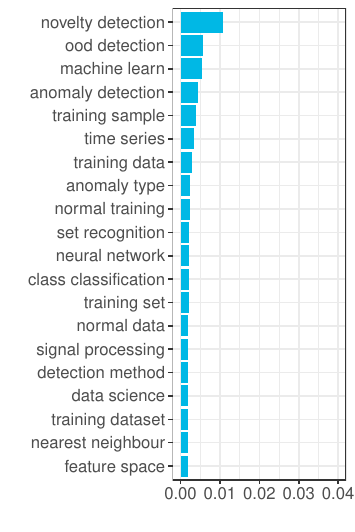}\hfill\mbox{}
  \caption{Most frequent 20 bigrams for topics 3 and 4.}
  \label{fig:lda_freq_3_4}
\end{figure}

\begin{figure}
  \centering
     \includegraphics{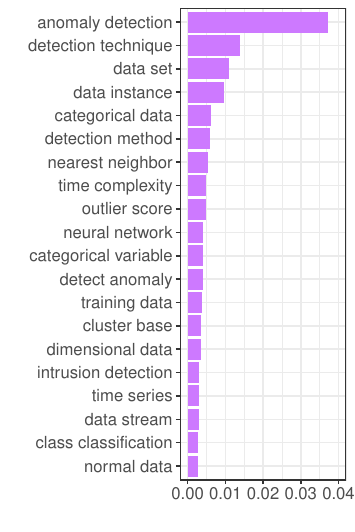}
  \caption{Most frequent 20 bigrams for topic 5.}
  \label{fig:lda_freq_5}
\end{figure}

The vocabulary used in the selected papers has been investigated using
a latent Dirichlet allocation (LDA) model \cite{Blei2003Latent}. The
model has been trained on the text extracted from the pdf files, after
some processing steps: suppression of special characters, numbers, and
stop words, and lemmatisation. LDA has been trained both on 1-grams
and 2-grams, and the number of topics has been selected using a
trade-off between different criteria
\cite{Griffiths2004Finding,Cao2009Density,Arun2010Finding,Deveaud2014Accurate}. The
model trained on 2-grams provided interesting results, with five
topics illustrated in Figure \ref{fig:lda_saliency} and an
associated clustering of the papers illustrated in Tables
\ref{tab:topic:1}, \ref{tab:topic:2}, \ref{tab:topic:3}, \ref{tab:topic:4}, and
\ref{tab:topic:4}. The clustering was obtained by using the topic
distribution of each paper as its numerical representation. We are
here in a particular case were each paper uses almost only one of the
four topics as thus the clustering is obvious: we assign each paper to
its dominating topic. Topics themselves are analysed using two
complementary illustrations: on the one hand, we extract top 0.1\% of
the most salient bigrams and represent their associated probabilities
in each topic in Figure \ref{fig:lda_saliency}, and on the other hand,
we represent the 20 most frequent terms in each topic in Figures
\ref{fig:lda_freq_1_2}, \ref{fig:lda_freq_3_4} and \ref{fig:lda_freq_5}.

Word saliency (bigrams in our case) has been introduced in
\cite{Chuang2012Termite} as a weighted Kullback-Leibler divergence
between the posterior distribution of the topics conditionally to a
specific word, and the marginal distribution of the topics. Hence,
Figure \ref{fig:lda_saliency} illustrates the most informative bigrams
in terms of how discriminant they are for the emerging topics. It may
be completed by Figure \ref{fig:lda_corrmat} in the Appendix, which
illustrates how these salient bigrams are similar to each other and
how blocks of meaningful content emerge within topics and within
documents.  

\begin{table}[ht]
\centering
\begin{tabular}{lrp{0.3\linewidth}}
  \toprule
Paper & year & Title \\ 
  \midrule
\small MarkouSingh2003NoveltyDetectionNeural \cite{MarkouSingh2003NoveltyDetectionNeural} & 2003 & Novelty detection: a review—part 2: neural network based approaches \\ 
\small MarkouSingh2003NoveltyDetectionStatistical \cite{MarkouSingh2003NoveltyDetectionStatistical} & 2003 & Novelty detection: a review—part 1: statistical approaches \\ 
\small SuTsai2011OutlierDetection \cite{SuTsai2011OutlierDetection} & 2011 & Outlier detection \\ 
\small ZimekSchubertEtAl2012SurveyUnsupervised \cite{ZimekSchubertEtAl2012SurveyUnsupervised} & 2012 & A survey on unsupervised outlier detection in high‐dimensional numerical data \\ 
\small Zhang2013AdvancementsOutlier \cite{Zhang2013AdvancementsOutlier} & 2013 & Advancements of outlier detection: A survey \\ 
\small ZimekFilzmoser2018ThereBack \cite{ZimekFilzmoser2018ThereBack} & 2018 & There and back again: Outlier detection between statistical reasoning and data mining algorithms \\
\small CarrenoInzaEtAl2020AnalyzingRare \cite{CarrenoInzaEtAl2020AnalyzingRare} & 2020 & Analyzing rare event, anomaly, novelty and outlier detection terms under the supervised classification framework \\ 
   \bottomrule
\end{tabular}
\caption{Papers assigned to topic 1} 
\label{tab:topic:1}
\end{table}

\begin{table}[ht]
\centering
\begin{tabular}{lrp{0.3\linewidth}}
  \toprule
Paper & year & Title \\ 
  \midrule
\small HodgeAustin2004SurveyOutlier \cite{HodgeAustin2004SurveyOutlier} & 2004 & A survey of outlier detection methodologies \\ 
\small AgyemangBarkerEtAl2006ComprehensiveSurvey \cite{AgyemangBarkerEtAl2006ComprehensiveSurvey} & 2006 & A comprehensive survey of numeric and symbolic outlier mining techniques \\ 
\small BoukercheZhengEtAl2020OutlierDetection \cite{BoukercheZhengEtAl2020OutlierDetection} & 2020 & Outlier detection: Methods, models, and classification \\
\small NassifTalibEtAl2021MachineLearning \cite{NassifTalibEtAl2021MachineLearning} & 2021 & Machine learning for anomaly detection: A systematic review \\ 
\small SamariyaThakkar2021ComprehensiveSurvey \cite{SamariyaThakkar2021ComprehensiveSurvey} & 2021 & A comprehensive survey of anomaly detection algorithms \\ 
   \bottomrule
\end{tabular}
\caption{Papers assigned to topic 2} 
\label{tab:topic:2}
\end{table}

\begin{table}[ht]
\centering
\begin{tabular}{lrp{0.3\linewidth}}
  \toprule
Paper & year & Title \\ 
  \midrule
\small Petrovskiy2003OutlierDetection \cite{Petrovskiy2003OutlierDetection} & 2003 & Outlier detection algorithms in data mining systems \\ 
\small BenGal2005OutlierDetection \cite{BenGal2005OutlierDetection} & 2005 & Outlier Detection in: Data Mining and Knowledge Discovery Handbook: A Complete Guide for Practitioners and Researchers \\ 
 \small HadiImon2009DectectionOutliers \cite{HadiImon2009DectectionOutliers} & 2009 & Detection of outliers \\ 
\small AguinisGottfredsonEtAl2013BestPractice \cite{AguinisGottfredsonEtAl2013BestPractice} & 2013 & Best-practice recommendations for defining, identifying, and handling outliers \\ 
 \small Rokhman2019SurveyMixed \cite{Rokhman2019SurveyMixed} & 2019 & A survey on mixed-attribute outlier detection methods \\ 
  \small WangBahEtAl2019ProgressOutlier \cite{WangBahEtAl2019ProgressOutlier} & 2019 & Progress in outlier detection techniques: A survey \\  
   \bottomrule
\end{tabular}
\caption{Papers assigned to topic 3} 
\label{tab:topic:3}
\end{table}

\begin{table}[ht]
\centering
\begin{tabular}{lrp{0.3\linewidth}}
  \toprule
Paper & year & Title \\ 
  \midrule
  \small PimentelCliftonEtAl2014ReviewNovelty \cite{PimentelCliftonEtAl2014ReviewNovelty} & 2014 & A review of novelty detection \\ 
    \small Foorthuis2021NatureTypes \cite{Foorthuis2021NatureTypes} & 2021 & On the nature and types of anomalies: A review of deviations in data \\ 
   \small SalehiMirzaeiEtAl2022UnifiedSurvey \cite{SalehiMirzaeiEtAl2022UnifiedSurvey} & 2022 & A unified survey on anomaly, novelty, open-set, and out-of-distribution detection: Solutions and future challenges \\ 
   \bottomrule
\end{tabular}
\caption{Papers assigned to topic 4} 
\label{tab:topic:4}
\end{table}

\begin{table}[ht]
\centering
\begin{tabular}{lrp{0.3\linewidth}}
  \toprule
Paper & year & Title \\ 
  \midrule
  \small ChandolaBanerjeeEtAl2009AnomalyDetection \cite{ChandolaBanerjeeEtAl2009AnomalyDetection} & 2009 & Anomaly detection: A survey \\ 
  \small TahaHadi2019AnomalyDetection \cite{TahaHadi2019AnomalyDetection} & 2019 & Anomaly detection methods for categorical data: A review \\ 
    \small ThudumuBranchEtAl2020ComprehensiveSurvey \cite{ThudumuBranchEtAl2020ComprehensiveSurvey} & 2020 & A comprehensive survey of anomaly detection techniques for high dimensional big data \\ 
  \small RuffKauffmannEtAl2021UnifyingReview \cite{RuffKauffmannEtAl2021UnifyingReview} & 2021 & A unifying review of deep and shallow anomaly detection \\ 
   \bottomrule
\end{tabular}
\caption{Papers assigned to topic 5} 
\label{tab:topic:5}
\end{table}

When studying Figures \ref{fig:lda_saliency}, \ref{fig:lda_freq_1_2},
\ref{fig:lda_freq_3_4} and \ref{fig:lda_freq_5}, 
it is worth noticing that topics appear to deal each with a
different point of view: the first speaks mainly of \emph{outlier detection} and \emph{novelty detection}, the second both of \emph{anomaly detection} and \emph{outlier detection}, the third essentially of \emph{outlier} detection, the fourth mostly of \emph{novely detection}, while the fifth mostly of \emph{anomaly detection.}

\begin{figure}
  \centering
  \includegraphics{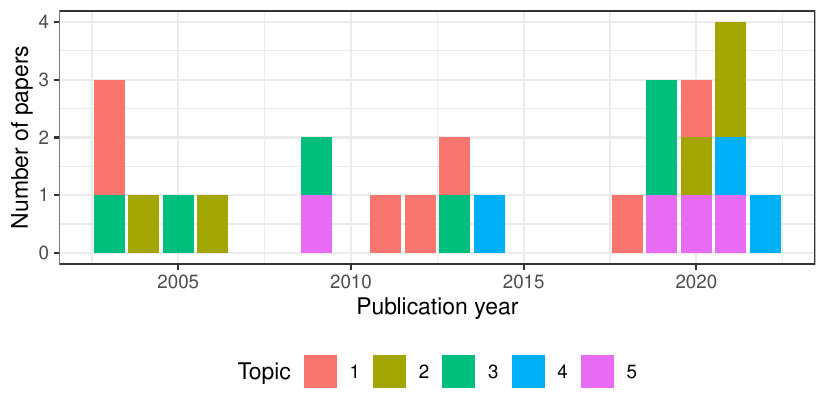}
  \caption{Topic distribution by publication year: each bar counts
    the number of papers published this year. The bar content is
    colored according to the topic to which is paper published this
    year is assigned.}
  \label{fig:topics_per_year}
\end{figure}

The papers have been
sorted according to publication time in table to emphasize the fact
that there is no obvious temporal structure in topics 1, 2 and 3,
while topics 4 and 5 could be qualified as more recent ones, even if
is contains the most cited survey
\cite{ChandolaBanerjeeEtAl2009AnomalyDetection} which is older than
the other papers. This weak temporal structure can 
be seen also on Figure \ref{fig:topics_per_year} and appears to some
extent in Figure  \ref{fig:papers_citations_mat} (papers of topic 4
are the less cited ones because they are somewhat recent). 

The first cluster of documents, associated to the first topic, contains the seven papers listed in Table \ref{tab:topic:1}. It contains the two historical surveys \cite{MarkouSingh2003NoveltyDetectionNeural}, \cite{MarkouSingh2003NoveltyDetectionStatistical}, as well as two papers by the same authors  \cite{ZimekSchubertEtAl2012SurveyUnsupervised}, \cite{ZimekFilzmoser2018ThereBack}. These surveys are aiming at bridging gaps between the statistical and the machine learning community, hence their similarity. Some terms such as \emph{rare event} or \emph{learn scenario} are extremely specific. Terms such as \emph{dimensional data}, or, to a lesser extent, \emph{neural network}, \emph{signal processing}, \emph{time series} or \emph{data stream} are also quite specific. For instance, the issue of high dimensional data is indeed addressed in detail in \cite{ZimekSchubertEtAl2012SurveyUnsupervised}, \cite{Zhang2013AdvancementsOutlier}, but also in \cite{SuTsai2011OutlierDetection}, and concern either the entire paper, or consistent sections of it.

The second topic alternates between \emph{outlier detection} and \emph{anomaly detection}, and is associated to the papers summarised in Table \ref{tab:topic:2}. These documents share common syntagms such as \emph{algorithm based}, \emph{detection algorithm}, \emph{mining techniques} or \emph{outlier mining}. \emph{Machine learning} is also specific to this topic, and at a closer look, several of the papers in this cluster are stemming from the data mining community \cite{AgyemangBarkerEtAl2006ComprehensiveSurvey} or the machine learning one \cite{NassifTalibEtAl2021MachineLearning}. The papers in this cluster provide quite general surveys, with very similar taxonomies, the later ones  inspired by the historical survey \cite{HodgeAustin2004SurveyOutlier}. 

The third topic contains most of the salient terms, while the papers associated to it and summarised in Table \ref{tab:topic:3} are speaking most specifically about \emph{outlier detection}.  On the one hand, this topic appears to focus more than the others on statistical approaches, hence the presence of bigrams such as \emph{model fit, parameter estimate, fit outlier, predict outlier, multilevel modelling ...}. Indeed, surveys such as   
\cite{BenGal2005OutlierDetection}, \cite{HadiImon2009DectectionOutliers} or \cite{WangBahEtAl2019ProgressOutlier}  propose taxonomies separating \emph{statistical methods} from \emph{machine learning} or \emph{data mining methods}. On the other hand, surveys in this topic, such as \cite{Rokhman2019SurveyMixed} are addressing the issue of data types, and particularly other features than numerical ones. Hence, bigrams such as \emph{numerical data, attribute data, mix attribute} appear with large frequencies. 

Topic four contains only three papers, listed in Table
\ref{tab:topic:4}, and very few salient terms: \emph{novelty
  detection} and \emph{ood detection}. It mixes papers that are quite
different and very specific: \cite{Foorthuis2021NatureTypes} discusses
anomaly types, \cite{PimentelCliftonEtAl2014ReviewNovelty} can be seen
as a modernized version of the companion papers
\cite{MarkouSingh2003NoveltyDetectionNeural} and
\cite{MarkouSingh2003NoveltyDetectionStatistical} as it leverages the
signal processing literature, while
\cite{SalehiMirzaeiEtAl2022UnifiedSurvey} is the only survey that
extends its scope to include subjects related to anomaly detection
such as out-of-distribution detection (hence the saliency of
\emph{ood detection}). 

The last topic concerns four documents listed in Table
\ref{tab:topic:5}, among which the historical
\cite{ChandolaBanerjeeEtAl2009AnomalyDetection} and the recent
and complete  \cite{RuffKauffmannEtAl2021UnifyingReview}. The other
two are a survey focusing on outliers for categorical data
\cite{TahaHadi2019AnomalyDetection}, or outliers for big data
\cite{ThudumuBranchEtAl2020ComprehensiveSurvey}. They all appear to
prefer using \emph{anomaly detection} instead of \emph{outlier} or
\emph{novelty detection}. In our opinion, all four are among the most
important surveys for the literature, whether by their historical and
wide spectrum value, by their unifying view, or by addressing specific
questions related to outlier detection. The salient bigrams for this
topic are \emph{categorical data} and \emph{categorical variable},
\emph{data set} or \emph{data instance}, or \emph{time complexity} and
\emph{dimensional data}.   

Overall this topic based analysis provides an idea of the variability
of the vocabulary used to describe outliers and anomalies, as well as,
the methods used to detect them. However, this variability seems to be
quite circumstantial and is somewhat explained by the pervasiveness of
outlier detection in many relatively separated fields such as signal
processing, statistics, data mining, etc. While this could be
interpreted as indicative of a fragmented field, we will see in the
next section that most of the surveys agree on numerous crucial
points.

\section{Consensual findings}\label{sec:consensus}
We discuss in this section the high level findings that appear to be
consensual throughout the selected surveys. 

\subsection{Definitions and taxonomies}\label{subsection:outlier:definition}

The quotes of \cite{BarnettLewis1978OutliersStatistical, Hawkings1980IdentificationOutliers} -discussed in Section \ref{sec:outliers-anomalies}- are often used \cite{ZhangMeratniaEtAl2007TaxonomyFramework, Aggarwal2017OutlierAnalysis, ZimekFilzmoser2018ThereBack} to illustrate the definition of outliers and most reviews agree that the definition is vague and application-dependent.
Chandola et al. write in \cite{ChandolaBanerjeeEtAl2009AnomalyDetection} ``\emph{Anomalies are patterns in data that do not conform to a well defined notion of normal behavior.}'' which constitutes already a step towards more formal definitions.

Some review propose a definition of their own as \cite{RuffKauffmannEtAl2021UnifyingReview}, \emph{an anomaly is an observation that deviates considerably from some concept of normality}.

It is only recently that the notion of outlier was defined in a precise mathematical sense in \citep{RuffKauffmannEtAl2021UnifyingReview}.

The literature is shared accross the fields of statistic, machine
learning, signal processing and data mining. Each field having its
application of interest and its own vocabulary. As most review insist
on, its leads to a wide literature with many words and expressions
covering related concepts from the classical outliers, anomaly,
novelty to the less frequent rogue values, mavericks, dirty data, etc.
In \cite{CarrenoInzaEtAl2020AnalyzingRare}, the authors tries to
disentangle the definitions of rare event, anomaly, novelty and
outlier. Hence, rare events are mostly found in problems of temporal
nature. Then, their distinction between the outlier, anomaly and
novelty is less consensual. According to their findings, outliers are
found in unsupervised scenario whereas novelty and anomaly are found
in supervised scenarios.  In
\cite{RuffKauffmannEtAl2021UnifyingReview}, Ruff et al. ma: papers of topics
4 and 5 tends to be less cited ke a subtle
and more consensual difference between anomaly (from a distinct
distribution), outlier (rare, low probability event) and novelty (instance
from a new region).

Historically, the underlying generating mechanism of outliers was used
to distinguish them from normal observations, as discussed in Section
\ref{sec:outliers-anomalies}: there were basically good observations
(with a heavy-tailed distribution for example) and contaminants.
\cite{ZhangMeratniaEtAl2007TaxonomyFramework}  distinguishes also
between two types of outliers that coincide with Hawkins' definition
\cite{Hawkings1980IdentificationOutliers}, errors and events
(generated by a different mechanism) to be identified for further
investigation. This tradition is still followed by recent surveys, for
instance by \cite{Foorthuis2021NatureTypes} for whom there is noise or actual signals.

This simple taxonomy is later enriched by \cite{AguinisGottfredsonEtAl2013BestPractice}, that proposes 5 main types of outliers (with a total of 13 subtypes). It covers the types of \cite{Hawkings1980IdentificationOutliers, ZhangMeratniaEtAl2007TaxonomyFramework} but also adds interesting types such as model outliers (having a large residual and possibly influencing the model) and cluster analysis outlier. 
This view on the model being at the center of outlier detection is later developed in \cite{ZimekFilzmoser2018ThereBack}. 
Some reviews also used the model either parametric, semi-parametric and non-parametric \cite{ZhangMeratniaEtAl2007TaxonomyFramework} or shallow and deep \cite{RuffKauffmannEtAl2021UnifyingReview} as an axis of analysis.

Influenced by the need to adapt to complex data, a more modern view
has emerged. The simpler case was opposing global outiers to local
outliers \cite{ZhangMeratniaEtAl2007TaxonomyFramework} (see Section
\ref{sec:local-versus-global} on this aspect).
Those definition are refined in \cite{ChandolaBanerjeeEtAl2009AnomalyDetection} that distingues point anomaly, contextual anomaly and collective anomaly. The simplest case is the one of \emph{point anomalies} where a single observation can be classified in isolation as an anomaly. This type of anomaly is the main focus of most of the methods. In statistical and machine learning terms, it is associated to the classical independence hypothesis between observations. 
When the observations are statistically dependent, the notion of anomaly should be revised. Indeed, the expected value of an observation is in this case dependent from the values of other observations. Then the status of an observation (normal or anomalous) cannot be decided in isolation. Such anomalies are called \emph{contextual anomalies} or \emph{conditional anomalies}. 
In \cite{Foorthuis2021NatureTypes}, only atomic and aggregated outliers are discussed (leaving aside the context).
\cite{RuffKauffmannEtAl2021UnifyingReview} adds to the types discussed
in \cite{ChandolaBanerjeeEtAl2009AnomalyDetection}  two more types
suited to the deep neural network case, low-level sensory anomaly and high-level semantic anomaly (which could be thought as subtypes of contextual anomalies). Low and high refer to the feature level in a deep learnin perspective. In images for example, a low-level sensory anomaly would be a the pixel or texture level whereas a high-level semantic anomaly would be the presence of an object in the image.

Data types were another axes of analysis of outliers. Originally, the main challenges were to extend the algorithms to multivariate cases and time-series \cite{BarnettLewis1978OutliersStatistical, Hawkings1980IdentificationOutliers,BeckmanCook1983Outliers}. Hence, a basic axis for a taxonomy would oppose univariate to multivariate cases.
It is now completed with more complex data type such as categorical
data \cite{TahaHadi2019AnomalyDetection,DivyaKumaran2016SurveyOutlier} and text, time-series and discrete sequences or spatial data \cite{ZhangMeratniaEtAl2007TaxonomyFramework}. This list is completed with graphs and networks in \cite{ChandolaBanerjeeEtAl2009AnomalyDetection,Aggarwal2017OutlierAnalysis}.
It must be noted that this discussion is completely absent of \cite{RuffKauffmannEtAl2021UnifyingReview} as they handle data through a feature map which could be adapted to complex data.

In summary, a consensus emerges from the literature on three axes of taxonomy. The outliers can be categorized according to
\begin{itemize}
\item the underlying \emph{generative mechanism}, characterizing outliers as errors, interesting or influential,
\item the underlying \emph{independence hypothesis}, leading to the definition of point, contextual and collective anomaly, 
\item the \emph{data type}, from simple univariate data to
  multivariate and structured data (including categorical data, times
  series, spatial data and graphs).
\end{itemize}
Globally, while taxonomies on anomaly types have evolved through time,
surveys tend to agree on the main separations (e.g. contextual versus
isolated). Many surveys do not even mention them owing to consensus
associated to the definition proposed in
\cite{ChandolaBanerjeeEtAl2009AnomalyDetection}. This is in stark
contrast with taxonomies on the methods themselves which are highly
variable as discussed in Section \ref{sec:taxonomies}.

\subsection{The high dimensionality issue}\label{sec:high-dimens-issue}
While extremely important, the impact of high dimensionality on
outlier detection performances is discussed in several reviews, but
not in the majority of them and not always by considering the effects
of the curse of dimensionality, beyond the complexity burden. For
example, many surveys considered in this study mention
\emph{distance-based} or \emph{density-based} methods, but only rarely
mention how data defined in a high-dimensional feature space may
impact - negatively - the performances of these methods.

Notice that any discussion on high dimensionality should distinguish
the dimension of the description space, that is the number of features
used to describe the entities under study, from the intrinsic
dimensionality of the data (see
e.g. \cite{DurrantKaban2009WhenNearest}). Outlier detection in the
description space is directly impacted by the \emph{curse of
  dimensionality} while methods that try first to reduce somehow the
dimensionality prior the application of a classical outlier detection
approach face this curse during the dimensionality reduction
phase. Following \cite{RuffKauffmannEtAl2021UnifyingReview}, it could
be argued that \emph{reconstruction models}, from principal component
analysis to auto-encoders, put their effort in the dimensionality
reduction phase, will density estimation models and one-class
approaches try to address directly the original data. In this section,
we discuss the way surveys present the difficulties of both
approaches.

One of the first surveys on unsupervised outlier detection,
\cite{ZhangMeratniaEtAl2007TaxonomyFramework} (unfortunately
unpublished) proposes a taxonomy where
the case of high-dimensional data sets is specifically considered, as
different from the \emph{simple data set} baseline. After recalling
that \emph{in high-dimensional spaces the data is sparse, the convex
  hull more difficult to discern and the notion of proximity less
  meaningful}, the survey focuses on subspace-based methods and on
some specific distance-based methods, and discusses the computational
complexity, the efficiency, and the difficulty to tune pre-defined
parameters for these two families of methods. 

The question of outlier detection in the context of high-dimensional
data has been studied in detail in
\cite{ZimekSchubertEtAl2012SurveyUnsupervised}. Several illustrations
allow to assess the main issues related to high-dimensionality and to
derive several consequences for outlier detection tasks: concentration
of scores, noise attributes, definition of reference sets, bias of
scores, interpretation and contrast of scores, exponential search
space, data-snooping bias, and eventually hubness. These problems
challenge the correctness of the methods, and the evaluation criteria
for assessing the validity of outlier detection. Traditional methods,
for instance, based on distance computations in the description space
and classified according to different taxonomies as
\emph{distance-based}, \emph{density-based}, \emph{nearest-neighbour
  based} or even \emph{clustering based} are thus generally severely
impacted by high-dimensionality. The rest of the paper discusses
methods suited for high-dimensionality, either from an
\emph{efficiency} or an \emph{effectiveness} point of view. Several
classes of methods - approximate neighbourhood computations, ensemble
methods, angle-based methods, subspace-based methods - are critically
discussed, while stressing the difficulties of actually evaluating the
different methods, particularly from a qualitative point of view.

The conclusion of \cite{ZimekSchubertEtAl2012SurveyUnsupervised} which
dates back to more than ten years ago states that
\begin{quote}
  \emph{the area of outlier detection specialised for high-dimensional
    data offers lots of opportunities for improvement. There are just
    a few approaches around in the literature so far, yet there are
    many directions to go and problems still to solve. The researcher
    should, though, be aware of the existing attempts of solution and
    the associated pitfalls.}
\end{quote}
With this in mind, it is at least surprising - and actually quite
troublesome - that many subsequent surveys do not discuss specifically
high-dimensionality issues, propose brief discussions lacking of
perspective, or, worse, continue to present \emph{distance-based} and
\emph{dissimilarity-based} methods without mentioning the curse of
dimensionality or considering it from the point of view of
computational burden only.  

The issues related to high-dimensionality and some of the problems
stressed in \cite{ZimekSchubertEtAl2012SurveyUnsupervised} have been
discussed however in some recent surveys, such as
\cite{WangBahEtAl2019ProgressOutlier},
\cite{ThudumuBranchEtAl2020ComprehensiveSurvey} and
\cite{BoukercheZhengEtAl2020OutlierDetection}. \cite{WangBahEtAl2019ProgressOutlier}
is mainly interested in the computational burden and misses a thorough
discussion on the effectiveness of the methods. Furthermore, the
taxonomy of the methods is not very helpful for assessing how they
deal - or not - with dimensionality
issues. \cite{BoukercheZhengEtAl2020OutlierDetection} picks up on
\cite{ZimekSchubertEtAl2012SurveyUnsupervised} and review more recent
methods, while preserving a taxonomy of methods according to their
efficiency or effectiveness in the high-dimensional
framework. \cite{ThudumuBranchEtAl2020ComprehensiveSurvey} also draw
extensively from \cite{ZimekSchubertEtAl2012SurveyUnsupervised}, by
recalling some of the issues implied by the curse of dimensionality,
and reviewing some subspace-based methods.  

Nevertheless, the question of high dimensionality appears to be still
an open one, and recent surveys are still quite void off thorough
discussions on the topic, beyond, as we mentioned, the computational
complexity. The fact that Aggarwal's text book
\cite{Aggarwal2017OutlierAnalysis} dedicates a simple chapter to the
issue and bases his discussion only on subspace methods is also
revealing: true high dimensional problems remain very difficult and
outlier detection is not anomalous with respect to this difficulty. A
possible explanation is the recent focus on deep learning approaches
which use very frequently a low dimensional latent representation as
shown in \cite{SalehiMirzaeiEtAl2022UnifiedSurvey} and are thus
targeting high dimensional issues via a form of intrinsic
dimensionality recovery.

\subsection{On the importance of anomaly scores}
Anomaly detection methods can output either a score that measures to
what extent an observation is anomalous or a binary label that
directly says whether the observation should be considered anomalous
or not. These two possible outcomes are at least mentioned by almost
all the surveys, starting early ones such as
\cite{HodgeAustin2004SurveyOutlier}. 

From a machine learning point of view, binary labeling is
attractive as it is a simple task: from a score it is always possible
to derive a classification via a simple thresholding while the reverse
is false. Thus labeling should reach better performances than scoring
considering the same resources (both in terms of data size and of
computational burden). Interestingly, while many survey papers cover
one-class approaches \cite{KhanMadden2014OneClass} the only one to
interpret it in terms of machine learning efficiency is
\cite{RuffKauffmannEtAl2021UnifyingReview}, under Vapnik's
simplicity principle. This is probably because many surveys miss the
fact that one-class methods are estimating a level set of the
probability density of the normal data, even if this aspect in only
implicit in the construction of the method.

While appealing on a machine learning point of view, labeling methods
are somewhat less convenient in practice. Firstly two recent surveys
emphasize the need for interpretable and explainable decisions
\cite{ZimekFilzmoser2018ThereBack,RuffKauffmannEtAl2021UnifyingReview}. According
to a recent survey on the subject
\cite{PanjeiGruenwalsEtAl2022SurveyOutlier}, scoring can be seen as a
form of minimal step in this direction.

In addition, scoring can be tuned \emph{a posteriori} to the
operational conditions: the scoring threshold between outliers and
normal data can be adapted to e.g. the human resources
available to investigate the anomalies. Using metrics such as the area
under the ROC curve (AUC) (or possibly better ones depending on the
trade-off between precision and false alarm rate, see
\cite{RuffKauffmannEtAl2021UnifyingReview}), one can evaluate the
performances of the scoring approach for the full range of decision
threshold. 

Finally, a score can be used to rank the instances rather than to
split them into outliers and normal ones. Notice however that
ranking is putting the weight of the decision on the analyst
shoulders, as they will have to decide where to stop in the list of
ranked observations. Even worse, the stopping decision could be driven
in this case by operational considerations in an opaque way (and even
possibly changing depending on external events). While those are valid
considerations, they should be explicitly stated. As stated in
\cite{ZimekFilzmoser2018ThereBack},
\begin{quote}
  \emph{At the end of the day the central question for any application
    of such outlier detection methods is how to statistically
    interpret the outlier score that has been provided by some
    method. This interpretation and its relationship to outlier scores
    of different methods is usually anything but obvious}.
\end{quote}
In summary, thresholding scores into decisions is part of the model
fitting process and should not be ignored. It is interesting to see
that while there is a consensus between the two surveys that discuss
to some extent the topic
\cite{ZimekFilzmoser2018ThereBack,RuffKauffmannEtAl2021UnifyingReview},
it is generally completely disregarded in the other surveys. 

\subsection{Learning conditions}\label{sec:learning:type}
It should be first pointed out that the learning conditions to be
discussed here are somewhat orthogonal to previously discussed scoring
or labeling as both can apply to either supervised and unsupervised
conditions (see \cite{ZimekFilzmoser2018ThereBack}).

There is a very clear consensus in the literature that anomaly
detection is done in majority in an unsupervised learning context as
far as the nature of the observations is concerned: the detection
algorithm works without knowing the true nature of the
observations. The recent survey
\cite{NassifTalibEtAl2021MachineLearning}, which uses a sound paper
collection methodology, reports that 58 \% of the papers it reviewed
can be associated to a specific learning paradigm. Among them, 46 \%
used an unsupervised learning paradigm. In almost all the surveys
included in the present paper, supervised methods occupy only a small
part of the discussion (for instance 4 pages out of 58 in
\cite{ChandolaBanerjeeEtAl2009AnomalyDetection}). This is also the
case in the main text book on the subject which dedicates only a
chapter to supervised models \cite{Aggarwal2017OutlierAnalysis}. 
Notable exceptions are \cite{SalehiMirzaeiEtAl2022UnifiedSurvey,CarrenoInzaEtAl2020AnalyzingRare}
discussed below.

However, the anomaly detection paradigm should not be confused with
the learning paradigm of the main task the analyst is trying to
solve. Indeed, as recalled in Section \ref{sec:estim-under-cont},
early statistical approaches treat outliers as a nuisance that impairs
estimation, in particular in a supervised context. For instance most
of \cite{BeckmanCook1983Outliers} is dedicated to the effect of
anomalies on the estimation of generalised linear models (see also
\cite{RousseeuwLeroy1987RobustRegression}). So the anomalous versus
normal nature of the examples is unknown, but the learning paradigm
could be supervised.

In the selected surveys, evocation of this subtlety seems to be
related to older papers with the exception of
\cite{AguinisGottfredsonEtAl2013BestPractice} (which is still in the
first phase of surveys, before 2018, see Section
\ref{sec:structure-field}). This is quite natural as the main
evolution of the field is to shift the attention from being robust to
outliers to detecting them. In addition, data mining oriented surveys,
such as \cite{ChandolaBanerjeeEtAl2009AnomalyDetection}, tend to
interpret robust methods from the point of view of outlier
detection. For instance edited linear regression techniques such as
Rousseeuw's \emph{least trimmed squares}
\cite{Rousseeuw1984LeastMedian} whose original aim is fitting a linear
model robustly in presence of outliers, is presented as a way to
detect outliers (by their large residuals). Statisticians are clearly
aware of this shift as exemplified in the surveys by
\cite{HadiImon2009DectectionOutliers} which dedicates a significant
space to statistical methods but explicitly restricts the discussion to
the detection setting in the unsupervised context. We refer the reader
to \cite{AguinisGottfredsonEtAl2013BestPractice} for thorough
discussion on the methodological aspects of the ``outliers as a
nuisance'' paradigm (albeit limited to the field of organizational
science).

Nevertheless, while quite uncommon, the case of supervised learning is
generally discussed in the surveys we selected. Indeed in some
application contexts such as fraud or computer intrusion detection, it
may be possible to collect a data set with labelled examples combining
(a lot of) normal examples and (a small set of) anomalous examples. In
this case, the problem is a standard but difficult supervised learning
one. The difficulties come from the imbalanced nature of the data
\cite{HeGarcia2009LearningImbalanced,Krawczyk2016LearningImbalanced,MaHe2013ImbalancedLearning},
as collecting examples of anomalous behaviour is generally difficult,
and from the ill-posed nature of the classification: while the normal
data class is well defined, the anomalous data form a collection of
unrelated examples that can exhibit vastly different
characteristics. In addition, the normal data set can be contaminated
by undetected anomalies \cite{RuffKauffmannEtAl2021UnifyingReview}.

Among the surveys studied here, only one is explicitly dedicated to
the supervised learning paradigm
\cite{CarrenoInzaEtAl2020AnalyzingRare}. It distinguishes anomaly
detection as the supervised case from outlier detection as the
unsupervised one, a distinction that we did not encounter elsewhere in
the literature. It also discusses variation over anomaly detection in
the context of rare event detection and novelty detection. We believe
that this survey is introducing an unfortunate confusion between
one-class learning \cite{KhanMadden2014OneClass} and supervised
learning, especially compared to the thorough discussion on the
subject in \cite{RuffKauffmannEtAl2021UnifyingReview} for instance,
but it has the merit of showing that anomaly detection is related
closely to other problems, especially in the supervised context. Those
problems are surveyed in \cite{SalehiMirzaeiEtAl2022UnifiedSurvey}
which extends the discussion to the more general setting of
out-of-distribution and open-set detection. This corresponds in
particular to situations where a supervised model is trained on a
subset of the classes it will be facing in the deployment phase.

Between those two extreme cases, with zero or full supervision,
different levels of partial supervision have been explored relatively
recently in the context of anomaly detection. While the concept is
briefly mentioned as early as in \cite{HodgeAustin2004SurveyOutlier},
it restricted to the idea of having examples of the normal data
(labelled as such) and a collection of unlabelled data, a framework
known as LPUE, Learning from positive and unlabeled
examples \cite{RuffKauffmannEtAl2021UnifyingReview}. More
importantly, early surveys do not discuss papers using weak
supervision beyond citing a few examples, even in surveys that
dedicate a (short) section to them
(e.g. \cite{SuTsai2011OutlierDetection}). A systematic coverage starts
only in relatively recent papers such as
\cite{WangBahEtAl2019ProgressOutlier} (and well as in
\cite{Aggarwal2017OutlierAnalysis}). The importance of semi-supervised
learning in a classical sense, i.e. when labels are also available for
the outliers, should not been understated as even a small number of
such labels can improve strongly the detection performances
\cite{RuffKauffmannEtAl2021UnifyingReview}. There is for this reason a
tendency to try and break out of the unsupervised setting by some form
of outlier ``generation'', especially in the deep learning community
\cite{SalehiMirzaeiEtAl2022UnifiedSurvey}. 

\subsection{Benchmarking}

Many of the reviews insist on the fact that benchmarks and open datasets are needed for the development of the field \cite{TahaHadi2019AnomalyDetection,CamposZimekEtAl2016EvaluationUnsupervised,ZimekFilzmoser2018ThereBack,RuffKauffmannEtAl2021UnifyingReview}. Indeed, repurposing classification datasets as it is done in some surveys may induce biases and limit the results and the conclusions. Using downsampling techniques in the evaluation procedures for building training and validation sets is also troublesome due to the scarcity and the large variance of the outliers. \cite{CamposZimekEtAl2016EvaluationUnsupervised} suggest that not only data, but also samples used for training should be made open and available for reproducibility purposes. 

It should be also mentioned that benchmarking is not difficult because of the lack of meaningful datasets only. The nature of the outliers themselves and the nature of the quantified outlierness by each of the methods make the task harder. The third aspect to consider is the lack of general and well understood evaluation metrics. Whereas AUC for instance appears to be plebiscited by the few surveys containing benchmarks, it has its own limits and biases \cite{RuffKauffmannEtAl2021UnifyingReview}. 

Beyond the issue of evaluation metrics for the effectiveness of outlier detection, the question of interpretability of the results stands out as at least as important. As mentioned in \cite{ZimekFilzmoser2018ThereBack}, \emph{the quest for a truly general and superior method is futile}, whereas \cite{RuffKauffmannEtAl2021UnifyingReview} recall that \emph{all models are wrong}. Since outlier detection is essentially an unsupervised task, benchmarking should be used for analysing and understanding the strengths and the weaknesses of each method, but also how and when certain models are wrong, especially when confronted to datasets with different characteristics. \cite{RuffKauffmannEtAl2021UnifyingReview} for example use methods neuralisation and apply explainable AI techniques to get some insights on the interpretability of the methods. 

Considering the above, one may wonder whether a general \emph{all-purpose} benchmarking is actually meaningful or useful. Taking into account the diversity in the nature of outliers, as stressed for instance in \cite{Foorthuis2021NatureTypes}, it would be rather uneasy to build general datasets including meaningful anomalies, and none of the reviews is actually providing any guidelines for building benchmarks of interest.    

\subsection{Neural network models}\label{sec:neur-netw-models}
A possibly surprising finding of our meta-survey is the prevalence
and staying power of artificial neural networks through this 20 years
period. The earliest survey considered here is the two parts one by
Markou and Singh
\cite{MarkouSingh2003NoveltyDetectionStatistical,MarkouSingh2003NoveltyDetectionNeural},
in which one part is entirely dedicated to neural
networks. \cite{Petrovskiy2003OutlierDetection} and latter
\cite{HodgeAustin2004SurveyOutlier} discuss also neural
networks. Auto-encoders are already popular as anomaly detection
techniques as exemplified by
\cite{SongShinEtAl2001AnalysisNovelty,HawkinsHeEtAl2002OutlierDetection}.

This early presence is followed by a reduced interest phase as papers in the
late 2000s and the early 2010s tend to mention neural networks only in passing or
not at all, as
\cite{AguinisGottfredsonEtAl2013BestPractice,HadiImon2009DectectionOutliers,SuTsai2011OutlierDetection,ZimekSchubertEtAl2012SurveyUnsupervised,Zhang2013AdvancementsOutlier}. As
already hypothesized in Section \ref{sec:structure-field}, this
is probably related to the decrease in popularity of neural networks
before the explosion of deep learning. In recent surveys, the absence
of neural networks seems to be related to the particular case of
categorical data
\cite{TahaHadi2019AnomalyDetection,Rokhman2019SurveyMixed}. \cite{Foorthuis2021NatureTypes}
is also a particular as it focuses only on anomaly types. By contrast,
the absence of neural network methods from
\cite{SamariyaThakkar2021ComprehensiveSurvey} is less easy to
interpret.

\begin{figure}
  \centering
  \includegraphics{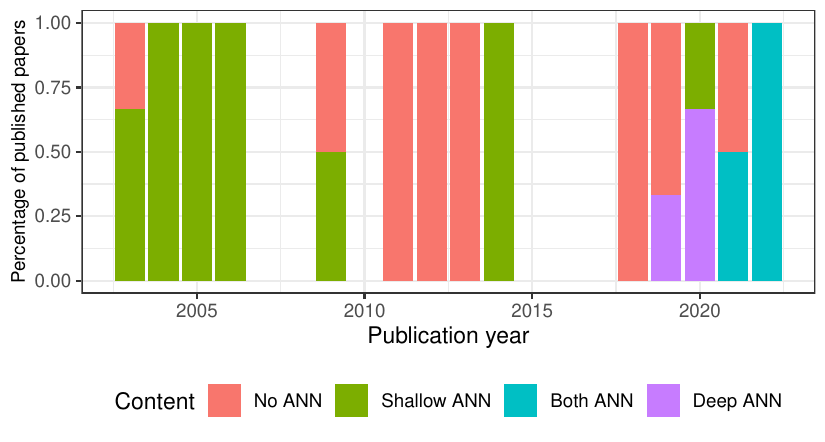}
  \caption{Inclusion of artificial neural network (ANN) papers per year:
    bars represent the distribution of the type of ANN papers
    considered by the surveys published during a given year.}
  \label{fig:ann_per_year}
\end{figure}

As expected, deep learning is present in all recent surveys. In
particular \cite{SalehiMirzaeiEtAl2022UnifiedSurvey} while it presents
itself as a generic survey could almost have been excluded from our
study as its discusses almost only deep learning approaches. This is
in stark contrast with \cite{RuffKauffmannEtAl2021UnifyingReview} which
goes beyond the separation between shallow and deep models to provide
a unifying view for learning an outlier detection model.

A synthetic representation of the evolution of coverage of neural
networks over the years is provided by Figure
\ref{fig:ann_per_year}. As explained above, we have a quite clear
decrease of interest in the early 2010s. The figure emphasizes another
interesting pattern: the resurgence of neural networks can be first
attributed to the appearance of deep learning methods (in 2019 and
2020), but recent papers tend to include again shallow artificial
neural networks.

As discussed in Section \ref{sec:specific-surveys}, deep learning is
popular enough to have generated 18 surveys on its use in outlier
detection (among them
\cite{ChalapathyChawla2019DeepLearning,PangShenEtAl2022DeepLearning}
are highly cited, despite the fact the first one is only an
unpublished technical report). Based on
\cite{SalehiMirzaeiEtAl2022UnifiedSurvey}, one can also argue that
deep learning allowed to unify problems related to outlier detection
(namely out-of-distribution and open-set detection) under a general
umbrella that could be described as the detection of non conformity of
the data to the hypothesized distribution assumption.

\section{Debated topics}\label{sec:debates}
We discuss in this section two selected topics for which the surveys
are not in agreement. 

\subsection{Local versus global}\label{sec:local-versus-global}
The idea that there are local and global outliers is quite popular in
the field and appeared relatively early
\cite{ZhangMeratniaEtAl2007TaxonomyFramework}, possibly as a
consequence of the introduction of the Local Outlier Factor (LOF) by
\cite{BreunigKriegelEtAl2000}. Other surveys mention this distinction,
mainly
\cite{WangBahEtAl2019ProgressOutlier,SamariyaThakkar2021ComprehensiveSurvey,BoukercheZhengEtAl2020OutlierDetection,ZimekFilzmoser2018ThereBack}.
The classical definition of local versus global anomalies, taken from
\cite{ZhangMeratniaEtAl2007TaxonomyFramework} is
\begin{quote}
  \emph{
A global outlier is an anomalous data point with
respect to all other points in the whole data set, but may not [be one] with respect to
points in its local neighborhood. A local outlier is a data point that is significantly
different with respect to other points in its local neighborhood, but may not be an
outlier in a global view of the data set.  }
\end{quote}
Obviously a global outlier must be a local one as a point 
anomalous with respect to \emph{all} the points in the whole data set,
is anomalous with respect to \emph{any} subset of those points. However the
definition introduces a paradox: if a point is
significantly different from its neighbors, i.e. by essence the points
that are the closest ones in the data set, then it must be
significantly different from all the other points. To resolve the
paradox, one must consider two different criteria: one criterion is
used to define a neighborhood and another one is used to characterize
local differences. This is done in the initial paper on LOF
\cite{BreunigKriegelEtAl2000} which considers k-nearest neighbors to
define the local subset of points and a local density estimator to
characterize each point.

More generally, as argued in \cite{SchubertZimekEtAl2014LocalOutlier},
the notion of ``local outliers'' should be refined in order to
distinguish the comparison scope and the characterization scope. The
\emph{comparison scope} associates to a given a point a subset of the
full data set and uses it as the basis of the decision to consider the
point as anomalous or not. The \emph{characterization scope} denotes
the subset of the data set used to build a characterization of a
point. For instance LOF is twice local.  Firstly, a point is only
compared to its neighbors. Secondly, each point is characterized by a
density computed on its neighbors. As shown in
\cite{SchubertZimekEtAl2014LocalOutlier} numerous methods are
considered \emph{local} but use in practice different combination of
local and global aspects (for instance a local comparison scope paired
with a global characterization scope).

While most of the literature focuses on the comparison scope,
\cite{RuffKauffmannEtAl2021UnifyingReview} argues that the locality
should refer to the characterization scope rather than to the
comparison scope. Using any complex \emph{global} density estimation
model (or one-class model), one can find small convex regions of low
density and thus identify outliers that would be miss by simpler
models, without the need for the estimation of a local model for each
data point. In other words, \emph{global} models can be used to detect
\emph{local} outliers. 

A parallel but somewhat similar discussion is provided in
\cite{Foorthuis2021NatureTypes} in which the opposition between local
and global outliers is rephrased in terms of contextual versus
non-contextual ones. This is also considered in
\cite{SchubertZimekEtAl2014LocalOutlier} which shows that another way
to circumvent the paradox of the local outlier definition is to use
some features of the points to define the comparison scope and the
rest of the features to define the characterization scope. A typical
example is given by outlier detection in time series, where time is
used to define the comparison scope and while the values of the
series are used to find possible outliers.

\subsection{On taxonomies}\label{sec:taxonomies}
As described in Section \ref{subsection:structure} most survey papers
in our selection use at least a classification of the methods they
discuss into several categories to organise their presentations. Many
of them arranged those categories into a hierarchical structure,
providing a taxonomy of outlier detection methods. One of the most
advanced of such taxonomies is proposed in
\cite{ZhangMeratniaEtAl2007TaxonomyFramework}. Perhaps unsurprisingly,
the consensus between those taxonomies and categorisations is
minimal.

We believe that this is a consequence of shoehorning a very diverse set
of methods into a collection of vaguely defined boxes. We already
cite \cite{BoukercheZhengEtAl2020OutlierDetection} which remarks that
\emph{distance based} and \emph{density based} outlier detection
methods are very frequently separated while they are all essentially
based on comparing distances to nearest neighbors. In some surveys,
finding the rationale of the categories is difficult: for instance
\cite{WangBahEtAl2019ProgressOutlier} as a \emph{learning based}
category from which clustering methods and ensemble methods are
excluded!

There are nevertheless some agreements. For instance, if we look past the
differences in names, some of the categories used in
\cite{PimentelCliftonEtAl2014ReviewNovelty} and
\cite{RuffKauffmannEtAl2021UnifyingReview} align nicely. They both use
a reconstruction based category and they somewhat agree on the
one-class/level set approaches (called \emph{domain based} in
\cite{PimentelCliftonEtAl2014ReviewNovelty}) and on the density
estimation methods (called \emph{probabilistic} in
\cite{PimentelCliftonEtAl2014ReviewNovelty}).

It seems to use that while the use of categories to organise the
presentation of collections of methods is almost mandatory,
identifying meaningful categories is a somewhat ill-posed and quite
difficult problem. Indeed we personally are convinced that the
categories proposed in \cite{RuffKauffmannEtAl2021UnifyingReview} (and
to some extent the ones in
\cite{PimentelCliftonEtAl2014ReviewNovelty}) are interesting because
they emphasize the main quality metric shared by the methods: density
estimation quality, level set quality or reconstruction quality. This
provides a high level view on the field and is somewhat orthogonal to
the implementation itself. However, this type of categorisation is
probably far less useful from a practical point of view. If a data
scientist is facing an outlier detection problem in e.g. the context
of fraud detection, how can they chose between a reconstruction error
approach or a density estimation one? 

\section{Conclusion}

At the end of the day, what does one draw from the reading of a large collection of surveys on anomaly and outlier detection? 

A very disappointing conclusion is both the large presence of
plagiarism and the almost total lack of paper collection methodology
even in high quality surveys. Both problems are probably not specific
to the field of outlier detection. Nevertheless is seems important to
emphasize that the paper collection bias is a well documented problem
in surveys and that the contribution of surveys to a field could only
improve by following standard procedures as outlined in
\cite{Kitchenham2004ProceduresPerforming} for instance. 

Those problems set apart, our meta-survey shows firstly that although
the literature is apparently extremely abundant, the number of surveys
actually contributing to the state of the art is rather limited, after
one has cleared the field. Secondly, it appeared to us that survey
approaches consisting in briefly summarising a list of methods and
proposing a - usually arbitrary - taxonomy are neither really useful
for the practitioners, nor meant to last overtime. In our opinion,
taxonomies centred around the anomaly types and/or the data
characteristics are more useful in practice and may be more easily
updated. Thirdly, we observed that important aspects such as the
computational complexity, the impact of high dimensionality, the
interpretability of the outlier score as a probability measure, or a
more unified view of the methods, should be more consistently discussed
when reviewing the state of the art. Nevertheless, only a small number
of surveys actually consider these issues, and bring perspective to
the field.

If one had to read one paper - or a couple of - to get a unified and
thorough view on anomaly detection, we suggest the recent survey
\cite{RuffKauffmannEtAl2021UnifyingReview}. Is is, to our knowledge,
the first to propose a formal mathematical definition of the notion of
outlier and to review a broad area of the field - including deep
neural networks - with a probabilistic perspective on the methods. An
alternative reading may be \cite{ZimekFilzmoser2018ThereBack} which
provides a high-level perspective on outlier detection, and attempts
to bring a statistical view on the methods, and a probabilistic
interpretation of the anomaly scores. Since the specific problems
related to the high dimensionality are not specifically addressed in
the two previous surveys, we also suggest
\cite{ZimekSchubertEtAl2012SurveyUnsupervised}, which provides the
most detailed discussion on the curse of dimensionality in the
framework of anomaly detection. To get a better understanding of the
diverse reality that is sometimes hidden under the generic concept of
outliers, we recommend the very detailed discussion on anomaly types
provided by \cite{Foorthuis2021NatureTypes}.  Eventually, Aggarwal's
monograph \cite{Aggarwal2017OutlierAnalysis} represents an important
reading, and contains a consolidated summary of the literature until
the mid 2010's.

Beyond those recommendations, we want to emphasize that numerous open
questions appearing in several of the surveys we selected should be
mentioned, such as the need for benchmark datasets and frameworks, and
the challenges related to the interpretability and the visualisation
of outliers. As it was discussed in a couple of surveys already, while
it is most probably useless to look for a universal effective method,
due to the different natures of outliers and to the unsupervised
framework in most situations, further investigation should be done in
the interpretability aspects.  

Finally, the role of artificial neural networks, and especially of
deep learning, in anomaly detection appears to be on the risen as
discussion in Sections \ref{sec:specific-surveys} and
\ref{sec:neur-netw-models}). The number of surveys we found without
specifically targeting them and the growing importance of those models
in recent generic surveys ask for a systematic review dedicated to
deep learning surveys. This could be done using a similar methodology
as the one used in the present paper. We would recommend to adapt the
search queries to the trends observed in our surveys, in particular in
terms of the generalized framework explored in
\cite{SalehiMirzaeiEtAl2022UnifiedSurvey}: one should not restrict the
search to outlier and anomaly, but rather expend it to include
expressions such as ``out-of-distribution'' and ``novelty''. Moreover,
the popularity of some particular type of deep models such as
Generative Adversarial Network (GANs) should be acknowledged: one
should not only search for surveys about ``deep learning'' but also
about GAN (see for instance \cite{XiaPanEtAl2022GanBased}) or probably
in the near future about Transformers
\cite{VaswaniShazeerEtAl2017AttentionAll}. 

\section*{Acknowledgments}
We thanks the two anonymous reviewers and the associated editor for
their insightful comments which help us to 
improve the quality of the manuscript.

\appendix
\section{Paper list}\label{appendix:paper-list}
\begin{center}\small
\begin{tabular}{lrrp{0.25\linewidth}}
  \toprule
Paper & Year & \multicolumn{1}{p{4em}}{Citations per year} & Title \\ 
  \midrule
MarkouSingh2003NoveltyDetectionNeural \cite{MarkouSingh2003NoveltyDetectionNeural} & 2003 & 55.45 & Novelty detection: a review—part 2: neural network based approaches \\ 
MarkouSingh2003NoveltyDetectionStatistical \cite{MarkouSingh2003NoveltyDetectionStatistical} & 2003 & 95.90 & Novelty detection: a review—part 1: statistical approaches \\ 
Petrovskiy2003OutlierDetection \cite{Petrovskiy2003OutlierDetection} & 2003 & 6.75 & Outlier detection algorithms in data mining systems \\ 
HodgeAustin2004SurveyOutlier \cite{HodgeAustin2004SurveyOutlier} & 2004 & 227.05 & A survey of outlier detection methodologies \\ 
BenGal2005OutlierDetection \cite{BenGal2005OutlierDetection} & 2005 & 2.39 & Outlier Detection \\ 
AgyemangBarkerEtAl2006ComprehensiveSurvey \cite{AgyemangBarkerEtAl2006ComprehensiveSurvey} & 2006 & 13.82 & A comprehensive survey of numeric and symbolic outlier mining techniques \\ 
ChandolaBanerjeeEtAl2009AnomalyDetection \cite{ChandolaBanerjeeEtAl2009AnomalyDetection} & 2009 & 868.64 & Anomaly detection: A survey \\ 
HadiImon2009DectectionOutliers \cite{HadiImon2009DectectionOutliers} & 2009 & 12.50 & Detection of outliers \\ 
SuTsai2011OutlierDetection \cite{SuTsai2011OutlierDetection} & 2011 & 4.83 & Outlier detection \\ 
ZimekSchubertEtAl2012SurveyUnsupervised \cite{ZimekSchubertEtAl2012SurveyUnsupervised} & 2012 & 77.00 & A survey on unsupervised outlier detection in high‐dimensional numerical data \\ 
AguinisGottfredsonEtAl2013BestPractice \cite{AguinisGottfredsonEtAl2013BestPractice} & 2013 & 116.50 & Best-practice recommendations for defining, identifying, and handling outliers \\ 
Zhang2013AdvancementsOutlier \cite{Zhang2013AdvancementsOutlier} & 2013 & 17.90 & Advancements of outlier detection: A survey \\ 
PimentelCliftonEtAl2014ReviewNovelty \cite{PimentelCliftonEtAl2014ReviewNovelty} & 2014 & 175.56 & A review of novelty detection \\ 
 \bottomrule
\end{tabular}
\end{center}

\begin{center}\small
\begin{tabular}{lrrp{0.25\linewidth}}
  \toprule
Paper & Year & \multicolumn{1}{p{4em}}{Citations per year} & Title \\ 
  \midrule
  ZimekFilzmoser2018ThereBack \cite{ZimekFilzmoser2018ThereBack} & 2018 & 28.80 & There and back again: Outlier detection between statistical reasoning and data mining algorithms \\ 
Rokhman2019SurveyMixed \cite{Rokhman2019SurveyMixed} & 2019 & 1.25 & A survey on mixed-attribute outlier detection methods \\ 
TahaHadi2019AnomalyDetection \cite{TahaHadi2019AnomalyDetection} & 2019 & 16.50 & Anomaly detection methods for categorical data: A review \\ 
WangBahEtAl2019ProgressOutlier \cite{WangBahEtAl2019ProgressOutlier} & 2019 & 70.25 & Progress in outlier detection techniques: A survey \\ 
BoukercheZhengEtAl2020OutlierDetection \cite{BoukercheZhengEtAl2020OutlierDetection} & 2020 & 49.33 & Outlier detection: Methods, models, and classification \\ 
CarrenoInzaEtAl2020AnalyzingRare \cite{CarrenoInzaEtAl2020AnalyzingRare} & 2020 & 17.33 & Analyzing rare event, anomaly, novelty and outlier detection terms under the supervised classification framework \\ 
ThudumuBranchEtAl2020ComprehensiveSurvey \cite{ThudumuBranchEtAl2020ComprehensiveSurvey} & 2020 & 44.00 & A comprehensive survey of anomaly detection techniques for high dimensional big data \\ 
Foorthuis2021NatureTypes \cite{Foorthuis2021NatureTypes} & 2021 & 15.50 & On the nature and types of anomalies: A review of deviations in data \\ 
NassifTalibEtAl2021MachineLearning \cite{NassifTalibEtAl2021MachineLearning} & 2021 & 28.50 & Machine learning for anomaly detection: A systematic review \\ 
RuffKauffmannEtAl2021UnifyingReview \cite{RuffKauffmannEtAl2021UnifyingReview} & 2021 & 192.50 & A unifying review of deep and shallow anomaly detection \\ 
SamariyaThakkar2021ComprehensiveSurvey \cite{SamariyaThakkar2021ComprehensiveSurvey} & 2021 & 5.50 & A comprehensive survey of anomaly detection algorithms \\ 
SalehiMirzaeiEtAl2022UnifiedSurvey \cite{SalehiMirzaeiEtAl2022UnifiedSurvey} & 2022 & 42.00 & A unified survey on anomaly, novelty, open-set, and out-of-distribution detection: Solutions and future challenges \\ 
 \bottomrule
\end{tabular}
\end{center}

\section{Additional PCA representation}\label{appendix:pca}
\begin{figure}
  \centering
  \includegraphics{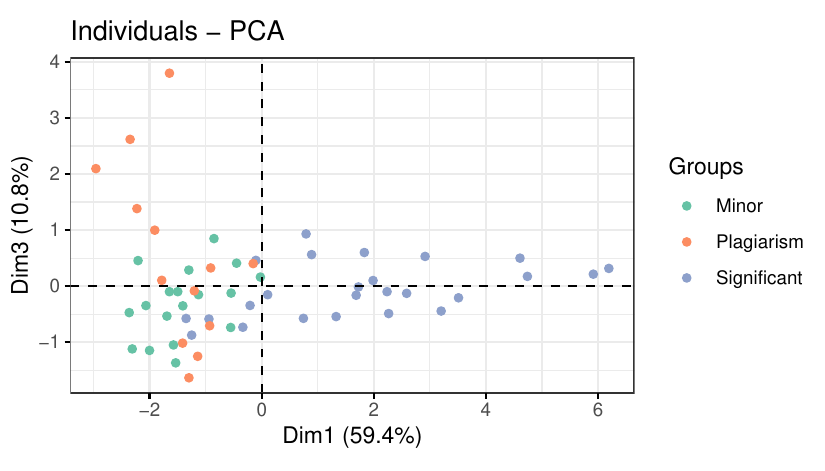}
  \caption{Principal component analysis results on the numerical
    characteristics of the survey papers: projection on the first and
  third principal components.}
  \label{fig:papers_pca_13}
\end{figure}

\begin{figure}
  \centering
  \includegraphics{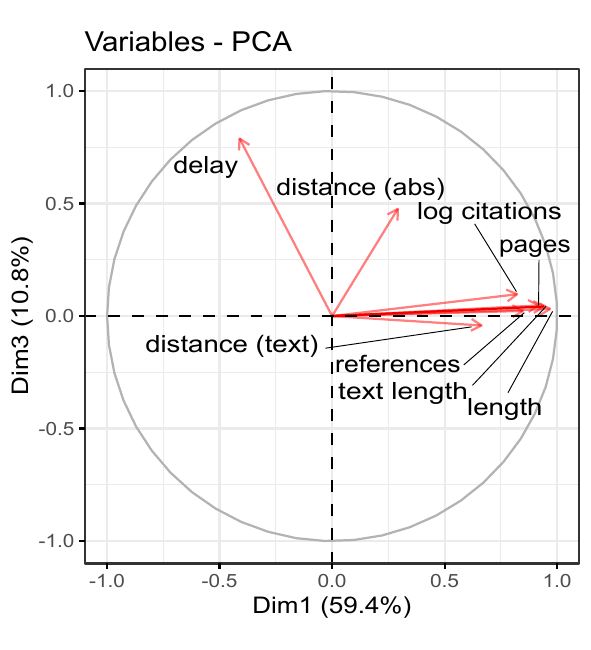}
  \caption{Contributions of the variables to the first and third principal
    components of the numerical characteristics of the survey papers.}
  \label{fig:papers_pca_var_13}
\end{figure}

The PCA results presented in Section \ref{sec:quality-assessment} can
be complemented by an analysis of the third component (motivated by
the scree plot on Figure \ref{fig:papers_scree}). As shown on Figures
\ref{fig:papers_pca_13} and \ref{fig:papers_pca_var_13}, the third
principal component (PC) plays a very similar role to the one of the second
component, bringing some separation between papers with plagiarism and
papers with minor contribution. While the second PC
tends to oppose distances to closest papers to delay between
publications and references, the third PC opposes the distances
computed on the full text to delays and distances computed on the
abstract. Distances play a stronger role in the second PC while the
third one focuses more on the delay. This shows more clearly that
plagiarism detection can be detected in some situation based on the
freshness of the references which cannot be hidden by a simple
rephrasing of the paper. Overall the third component confirms that our
selection process aligns with simple characteristics of the papers. 

\section{Additional LDA illustrations}\label{appendix:lda}

\begin{figure}
  \centering
  \includegraphics{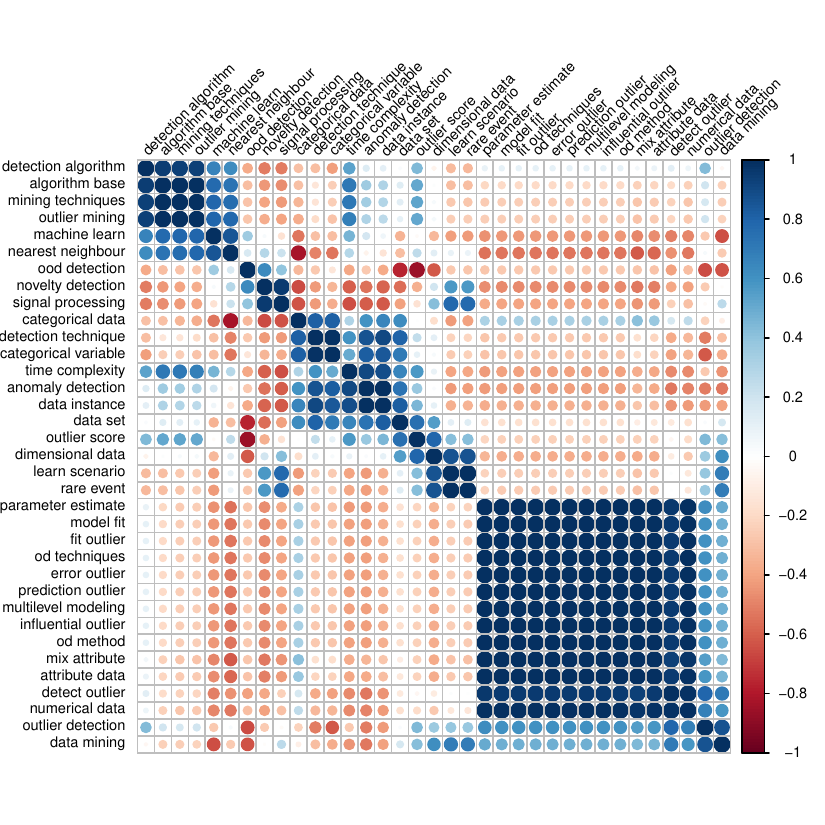}
  \caption{Correlation matrix of the most salient 0.1\% bigrams, according to the LDA model. Bigrams are ordered according to a hierarchical clustering.}
  \label{fig:lda_corrmat}
\end{figure}

Figure \ref{fig:lda_corrmat} contains an illustration of the correlation matrix computed for the most salient 0.1\% bigrams in the final corpus, and according to the LDA model outputs. Correlations are measuring the cosine-similarity on the bigram profiles, as given by their relative frequencies within each topic. Within the matrix, bigrams are ordered according to a hierarchical clustering. As one may easily see, bigrams are naturally grouped into blocks with similar contents. 

\bibliographystyle{elsarticle-harv}
\bibliography{biblio_C3,biblio_C5,biblio_C6,biblio_benchmarks,biblio_historical,
biblio_arxiv,biblio_other}

\end{document}